\documentclass[11pt]{article}
\usepackage{geometry}

\usepackage{authblk}
\usepackage{graphicx} 
\usepackage{amsmath} 
\usepackage{amssymb} 
\usepackage{cite} 
\usepackage{booktabs}
\usepackage{hhline}
\usepackage{float}
\usepackage{changepage} 
\usepackage{tabularx} 
\usepackage{xcolor}
\usepackage{soul}
\sethlcolor{yellow}
\usepackage{bm}
\usepackage[ruled,vlined]{algorithm2e}
\usepackage{tikz}
\usepackage{upgreek}
\usepackage{eurosym}

\title{Closed-Loop CO$_2$ Storage Control With History-Based Reinforcement Learning and Latent Model-Based Adaptation}

\author{Sofianos Panagiotis Fotias $^{1,*}$, Vassilis Gaganis $^{1,2}$ }

\begin{document}

\maketitle
\begin{center}
 \begin{tabular}{l}
 \textsuperscript{1}School of Mining and Metallurgical Engineering, National Technical University\\ of Athens, Athens 157 73, Greece \\
 \textsuperscript{2}Institute of Geoenergy, Foundation for Research and Technology, \\Chania, 73100, Greece \\
 \thanks{$^*$Correspondence: sfotias@metal.ntua.gr}

 \end{tabular}
\end{center}


\begin{abstract}

Closed-loop management of geological CO$_2$ storage requires control policies that adapt to uncertain reservoir behavior while relying on observations that are realistically available during operation. This work formulates CO$_2$ injection and brine-production control as a partially observable sequential decision problem and studies deployable deep reinforcement-learning controllers trained with high-fidelity reservoir simulation. We first compare privileged-state, well-only, history-conditioned, masking-curriculum, and asymmetric teacher-student model-free policies in order to quantify the value of temporal well-response information and training-time privileged simulator states. We then evaluate a latent model-based adaptation pipeline that reuses nominal latent dynamics and retunes controllers under known injector failure, leakage-induced dynamics and reward shift, and compartmentalized reservoir connectivity. The results show that history-conditioned policies recover nearly all of the privileged-state performance while using only deployable well-level information, and that latent model-based retuning outperforms direct model-free retuning under the same scenario-specific real-simulator budget in the abnormal operating cases. The proposed framework therefore provides a simulator-budget-aware alternative to repeated online history matching and re-optimization for closed-loop CO$_2$ storage control.

\end{abstract}


\section{Introduction}
Achieving net-zero emissions by mid-century requires rapid deployment of near-zero carbon technologies across sectors \cite{jarvis2018technologies}. Carbon Capture, Utilization and Storage (CCUS) is widely considered a key option because it can mitigate emissions from hard-to-abate industries and enable deep reductions where alternatives are limited \cite{gabrielli2020role,bui2021role}. In geological storage, captured CO$_2$ is injected into subsurface formations for permanent containment; suitable sites must provide adequate pore volume and injectivity, alongside a competent seal to prevent unwanted migration \cite{rackley2017introduction,tomic2018criteria}. Saline aquifers and depleted hydrocarbon fields are among the most mature storage candidates, supported by extensive project experience and monitoring data \cite{ji2015co2,bachu2015review,michael2010geological,hannis2017co2,mohammadian2019co,li2006co2}. Despite this maturity, storage performance and containment remain strongly influenced by uncertain geology and nonlinear multiphase flow, motivating decision-making frameworks that can adapt as information becomes available.
\\\\
Operational CCS planning and management rely heavily on numerical simulation to evaluate injectivity, pressure buildup, plume evolution, and trapping mechanisms over the project life cycle \cite{ismail2023carbon,pruess2004code,class2009benchmark}. Building on simulation-based forecasting, optimizing the operational schedule is essential for maximizing sequestration while meeting constraints and regulatory guidelines \cite{cihan2015optimal,cameron2012optimization,bachu2008co2,de2009acceptability}. In practice, CCS strategies often aim to sustain high CO$_2$ injection while regulating pressure buildup through brine production; producers may be shut in after significant CO$_2$ breakthrough, and operations are typically curtailed when bottomhole pressure approaches safety limits \cite{bandilla2017active,buscheck2016pre,anderson2020estimating}. In open systems with stratigraphic traps, pressure can also be relieved through brine migration to neighboring formations \cite{wang2012investigation}. These operational choices define the main decision variables for subsurface management. In the present work, we focus on time-varying CO$_2$ injection and brine-production controls. Common objectives include maximizing stored CO$_2$, improving trapping quality (e.g., immobile-to-mobile ratios), and economic measures such as NPV, while respecting pressure and containment constraints. Since each candidate control schedule requires a fully implicit simulation, the resulting objectives are expensive, nonlinear, and naturally treated as black-box functions in constrained multi-objective optimization \cite{deb2016multi}.
\\\\
Classic reservoir optimization is often conducted in an open-loop manner, where a control schedule is optimized against a fixed digital twin of the subsurface \cite{islam2020holistic}. This is limiting in CCS because the true system response is uncertain before operations begin, and decisions should ideally incorporate newly acquired monitoring data over time. Open-loop approaches can partially address uncertainty by optimizing over ensembles of geological realizations, but this increases computational cost substantially and can drive overly conservative solutions when constraints must hold across all models. Surrogate modeling and Bayesian optimization have been widely adopted to reduce simulator calls and quantify uncertainty in expensive black-box objectives \cite{cozad2014learning,frazier2018tutorial,adler1990introduction}, including explicit distinctions between aleatoric and epistemic uncertainty \cite{der2009aleatory,hullermeier2021aleatoric}. Nevertheless, the sequential nature of CCS operations motivates a closed-loop viewpoint in which controls are updated as new information arrives. Our previous work explored uncertainty-aware open-loop optimization for CCS through Bayesian optimization with permutation-invariant Gaussian-process surrogates \cite{fotias2024optimization}. The present study moves from that pre-operational setting to sequential closed-loop decision making.
\\\\

Closed-loop reservoir management (CLRM) addresses sequential subsurface decision-making by assimilating newly acquired data, updating geological models, and re-optimizing controls over time \cite{brouwer2004improved,aitokhuehi2005optimizing,sarma2006efficient,jansen2009closed,wang2009production,bukshtynov2015comprehensive}. Its practical use in CCS, however, is limited by the repeated cost of history matching and simulation-based optimization, especially when large ensembles of plausible geological models must be maintained \cite{dadashpour2010derivative,liu2020multilevel,chen2006data,emerick2013ensemble}. In this work, we formulate closed-loop CO$_2$ storage control as a partially observable sequential decision problem. Policies are trained offline on prior digital-twin realizations and evaluated on a held-out target reservoir under information regimes that distinguish between privileged simulator fields and deployable well-level observations \cite{zhang2022training,nasir2023deep,parisotto2020stabilizing}.  The first part of the paper compares model-free SAC controllers across observability regimes, from privileged-state access to well-history-based deployment. The second part introduces a latent model-based adaptation workflow for abnormal operating conditions, including injector-control loss, leakage-induced dynamics and reward shift, and compartmentalized connectivity. All abnormal-scenario comparisons are reported under explicit real-simulator interaction budgets.

\section{Related Work}
\subsection{Closed-loop reservoir management (CLRM)}
CLRM frameworks have been developed to enable sequential decision-making by repeatedly assimilating production data (history matching) and re-optimizing operational controls \cite{brouwer2004improved,aitokhuehi2005optimizing,sarma2006efficient,jansen2009closed,wang2009production,bukshtynov2015comprehensive}. Within CLRM, production optimization has been addressed using derivative-free methods \cite{bouzarkouna2012well,isebor2014derivative,nasir2020hybrid,onwunalu2010application}, gradient-based approaches \cite{alhuthali2008optimal,liu2020sequential,wang2002optimization}, and ensemble-based formulations \cite{fonseca2015ensemble,fonseca2015improving,ramaswamy2020improved}. A persistent challenge is that history matching is itself a difficult inverse problem \cite{dadashpour2010derivative,liu2020multilevel,chen2006data,emerick2013ensemble} and must be repeated throughout operations, increasing the computational burden and expanding the ensemble of plausible models. A related extension is closed-loop field development (CLFD), where drilling decisions are optimized jointly with operational control \cite{bellout2012joint,li2013simultaneous,forouzanfar2014joint}.

\subsection{Representative model selection and uncertainty propagation in CLRM}
Because CLRM can produce large ensembles of updated models, several works have focused on selecting representative subsets that preserve key flow responses while reducing simulation cost. Shirangi and Durlofsky proposed low-dimensional representations that combine permeability-based and flow-based descriptors and studied subset-selection accuracy using differences in flow response \cite{shirangi2016general}. They further introduced closed-loop optimization with sample validation, where the representative set is adaptively expanded if a validation criterion is not met \cite{shirangi2015closed}. For uncertainty propagation across multiple stages of reservoir modeling and development, Arnold et al. and Hutahaean et al. used neighborhood sampling methods (NAB) to sample posterior distributions of history-matched models, enabling uncertainty quantification for subsequent optimization steps \cite{arnold2016optimisation,hutahaean2019reservoir,sambridge1999geophysical1,sambridge1999geophysical2}.

\subsection{Deep reinforcement learning for reservoir control}
Deep reinforcement learning has been increasingly explored as an alternative to repeated online optimization by learning policies that map observations directly to control actions. Early work applied value-based and on-policy methods to subsurface control problems, including SARSA for SAGD \cite{rummery1994line} and Q-learning-based approaches for waterflooding \cite{watkins1992q,hourfar2019reinforcement}. Subsequent studies compared modern DRL algorithms against heuristic optimizers and investigated training across multiple geological realizations to account for uncertainty \cite{ma2019waterflooding}. Policy-gradient methods have also been used with high-dimensional state representations (e.g., pressure/saturation images) \cite{miftakhov2020deep,schulman2017proximal}. 

More CLRM-aligned workflows train agents on sets of prior models selected to represent geological uncertainty, and compare DRL against evolutionary strategies \cite{dixit2022stochastic,mnih2016asynchronous}. Soft Actor-Critic (SAC) has become a common choice for continuous well-control problems \cite{haarnoja2018soft}, and convolutional architectures have been shown to be advantageous when dense spatial reservoir states are available as inputs \cite{zhang2022training}. Related representation-learning pipelines embed reservoir states into latent spaces (e.g., via VAEs) to enable policy learning and transfer to unseen but similar environments \cite{wang2022deep}. Hybrid evolutionary-RL frameworks have also been proposed to improve policy diversity and robustness \cite{wang2023evolutionary}.

\subsection{Partial observability and sequence models for CLRM-like control}
In practical CCS/CLRM settings, the control problem is naturally partially observable: operators primarily observe well measurements and sparse monitoring rather than full-field states. This motivates POMDP formulations and history-dependent policies that integrate temporal context. Nasir et al. explicitly formulated CLRM as a POMDP and introduced sequence-based policies using gated transformer blocks (GTrXL) to process observation histories, demonstrating improved control under partial observability relative to baseline CLRM strategies \cite{nasir2023deep,parisotto2020stabilizing}. Their subsequent work incorporated practical operational considerations through action and reward shaping and extended the approach to multi-asset settings \cite{nasir2023practical,nasir2024multi}. DRL has also been applied in closed-loop field development (CLFD), where sequential drilling decisions are optimized jointly with controls \cite{he2022deep,nasir2021deep}.

\section{Problem Formulation and Model-Free Methodology}

\subsection{Sequential CCS control problem}
\begin{figure}[t]
	\centering
	\includegraphics[width=0.9\linewidth]{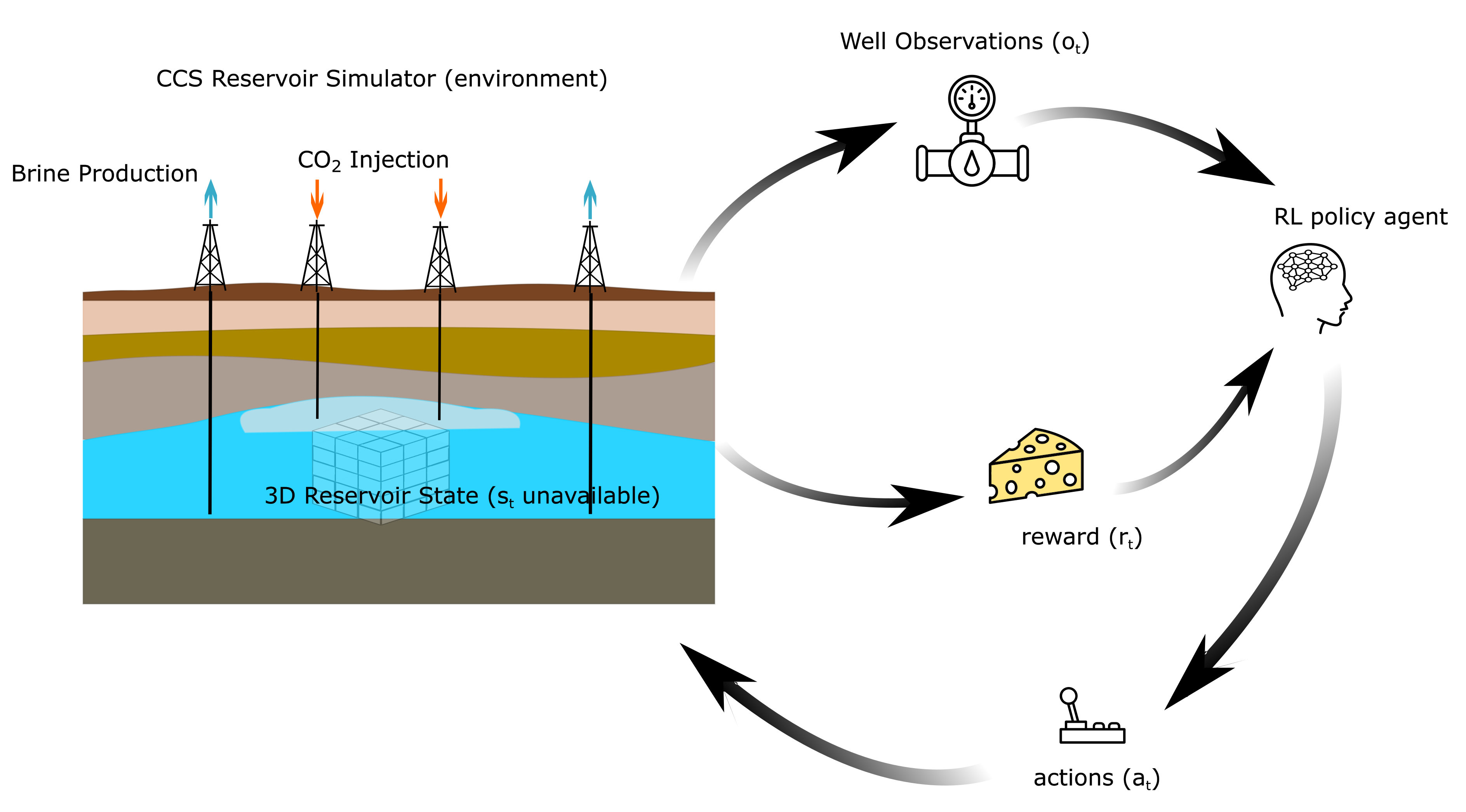}
	\caption{Closed-loop CCS control formulation considered in this work. The reservoir simulator evolves the latent subsurface state under the applied actions, while the agent receives only well-level observations and reward signals. The full spatial reservoir state remains latent to the deployed controller.}
	\label{fig:clrm_pomdp}
\end{figure}

We consider closed-loop reservoir management for geological CO$_2$ storage as a finite-horizon sequential decision-making problem. At each control step, the agent selects injection and production controls, the reservoir simulator advances the system dynamics, and new observations are returned for the next decision. The objective is to maximize storage-related operational performance while satisfying the physical and operational limits of the reservoir system.

In the present study, the action vector is continuous and contains eleven control variables,
\begin{equation}
a_t \in \mathbb{R}^{11},
\end{equation}
corresponding to the rate settings of eight producer wells and three CO$_2$ injector wells. The policy operates in a normalized action space, and each action component is subsequently mapped to its corresponding physical range before being written to the simulation deck. Producer controls are mapped to admissible production rate intervals, while injector controls are mapped to admissible gas injection rate intervals. This normalization allows all model variants to share the same action parameterization and learning setup.

Each decision is applied over one control interval of two years, implemented in the simulator through a time step of 730 days. An episode consists of 20 such control updates, giving rise to a long-horizon closed-loop management problem. After each control update, the simulator outputs are used both to construct the next observation and to evaluate the reward.

The reward function is designed to reflect the main operational trade-offs in CO$_2$ storage. At each decision step, we evaluate a storage-related cash-flow surrogate that rewards retained injected CO$_2$ while penalizing gas production and brine handling. In addition, a small bonus is assigned when the net storage rate remains within a desired operating window, emulating realistic CCS project needs. Denoting by $r_t$ the resulting scalar reward, the learning objective is to maximize the expected return
\begin{equation}
J(\pi) = \mathbb{E}_{\pi}\left[\sum_{t=0}^{T-1} r_t \right],
\end{equation}
where $T=20$ is the episode length and the expectation is taken over simulator transitions and, when applicable, reservoir-model uncertainty. Table~\ref{tab:reward_constraints} summarizes the reward and constraint components used by the environment. Figure~\ref{fig:clrm_pomdp} provides a schematic overview of the sequential CCS control loop, including the interaction between controls, simulator evolution, and available observations.

\begin{table}[t]
	\centering
	\caption{ Reward and constraint components used by the simulator environment. The same base reward is used across the nominal settings; Scenario~2 adds the leakage penalty.}
	\label{tab:reward_constraints}
	\small
	\begin{tabularx}{\textwidth}{p{0.23\textwidth}X p{0.18\textwidth}}
		\toprule
		Component & Implementation role & Used in \\
		\midrule
		Retained CO$_2$ value & Present-value reward based on injected minus produced gas & All scenarios \\
		Produced-gas penalty & Penalizes CO$_2$ production/breakthrough through the cash-flow term & All scenarios \\
		Brine-handling penalty & Penalizes produced brine treatment cost & All scenarios \\
		Net-storage operating bonus & Adds a small bonus when sequestration rate lies in the target interval & All scenarios \\
		Leakage penalty & Penalizes gas saturation above threshold in the leakage region & Scenario~2 \\
		Action bounds & Maps normalized actions to admissible producer/injector rate ranges & All scenarios \\
		Simulator/deck limits & Enforces well and operational limits during OPM/Flow execution & All scenarios \\
		\bottomrule
	\end{tabularx}
	
\end{table}

\subsection{POMDP formulation}
The CCS control problem is naturally formulated as a partially observable Markov decision process (POMDP),
\begin{equation}
\mathcal{P} = (\mathcal{S}, \mathcal{A}, \mathcal{O}, P, \Omega, R, \gamma),
\end{equation}
where $s_t \in \mathcal{S}$ denotes the latent reservoir state, $a_t \in \mathcal{A}$ the control action, $o_t \in \mathcal{O}$ the observation available to the agent, $P(s_{t+1}\mid s_t,a_t)$ the simulator-induced transition kernel, $\Omega(o_t\mid s_t)$ the observation model, $R(s_t,a_t)$ the reward function, and $\gamma$ the discount factor used by the reinforcement-learning algorithm.

\begin{figure}[t]
    \centering
    \includegraphics[width=0.92\linewidth]{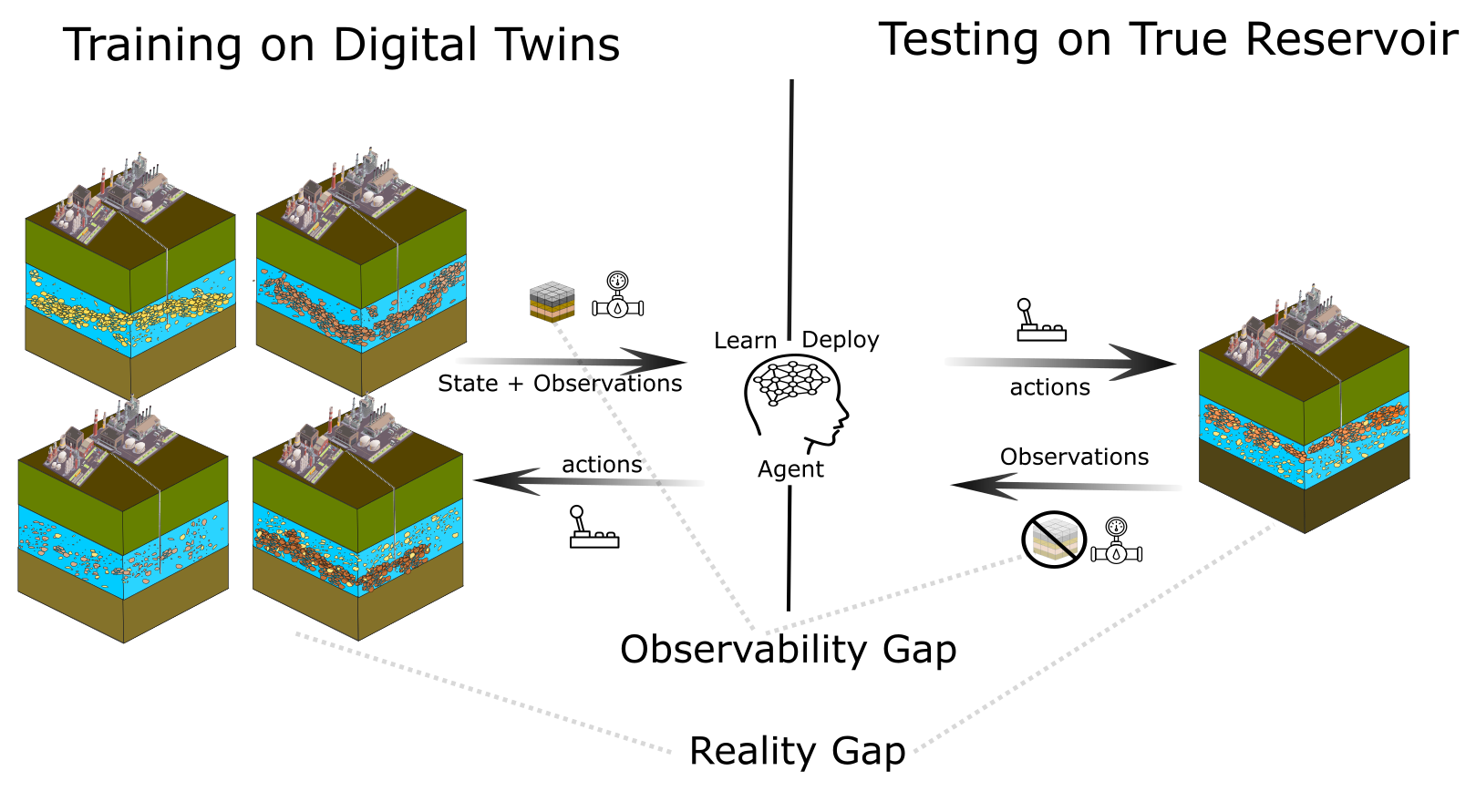}
    \caption{Training-deployment setting considered in this work. Policies are trained on ensembles of prior digital twins, whereas deployment takes place on the true reservoir using only practical field observations. This creates both a \emph{reality gap}, due to mismatch between training models and the true system, and an \emph{observability gap}, because dense simulator states available during training are not available during real operation.}
    \label{fig:uncertainty_gap}
\end{figure}

The latent state contains the full physical information required to evolve the system under the numerical simulator, including pressure and saturation distributions together with the effect of the applied controls. In contrast, the agent does not necessarily observe this full state. In practical closed-loop operation, decisions must be made from sparse monitoring information, primarily well responses and limited field measurements, rather than from complete spatial flow fields. This distinction is central to the present work. To make this explicit, we define the policy input through an observation map
\begin{equation}
o_t = h(s_t),
\end{equation}
and study multiple choices of $h(\cdot)$ corresponding to the information regimes defined below. Across all model-free variants, the action space, simulator dynamics, and reward definition are unchanged; only the policy information set differs. This isolates the effect of observability while retaining the same control task. Because training uses prior digital twins and evaluation uses the target reservoir, the experiments also include a train-test reality gap. Figure~\ref{fig:uncertainty_gap} summarizes the resulting observability and reality gaps.

\subsection{Baseline extremes of observability}

We use two endpoint baselines to define the observability range. The privileged-state benchmark is a diagnostic upper reference that uses dense target-reservoir simulator fields together with well observations. The well-only baseline is the most restrictive deployable controller and uses only the current well-observation tensor during training on prior digital twins and evaluation on the target reservoir. The later variants retain the deployable observation interface but add either temporal context or asymmetric privileged training signals.

\begin{table}[t]
	\centering
	\caption{Model-free information regimes considered in Section~3.}
	\label{tab:model_free_information_regimes}
	\small
	\begin{tabularx}{\textwidth}{p{0.21\textwidth}p{0.23\textwidth}p{0.27\textwidth}X}
		\toprule
		Variant & Policy input & Privileged training signal & Role \\
		\midrule
		Privileged-state & Spatial fields + wells & Policy observes dense target state & Oracle upper reference \\
		Well-only & Current well tensor & None & Sparse deployable baseline \\
		History-conditioned & Well history + current wells & None & Tests temporal context \\
		Masked-critic curriculum & Current wells & Critic receives masked spatial state & Tests critic-side privileged dependence \\
		Teacher-student & Well history + current wells & Teacher critics / distillation & Privileged value supervision \\
		\bottomrule
	\end{tabularx}
\end{table}

\subsubsection{Privileged-state benchmark}
\begin{figure}[t]
	\centering
	\includegraphics[width=0.85\linewidth]{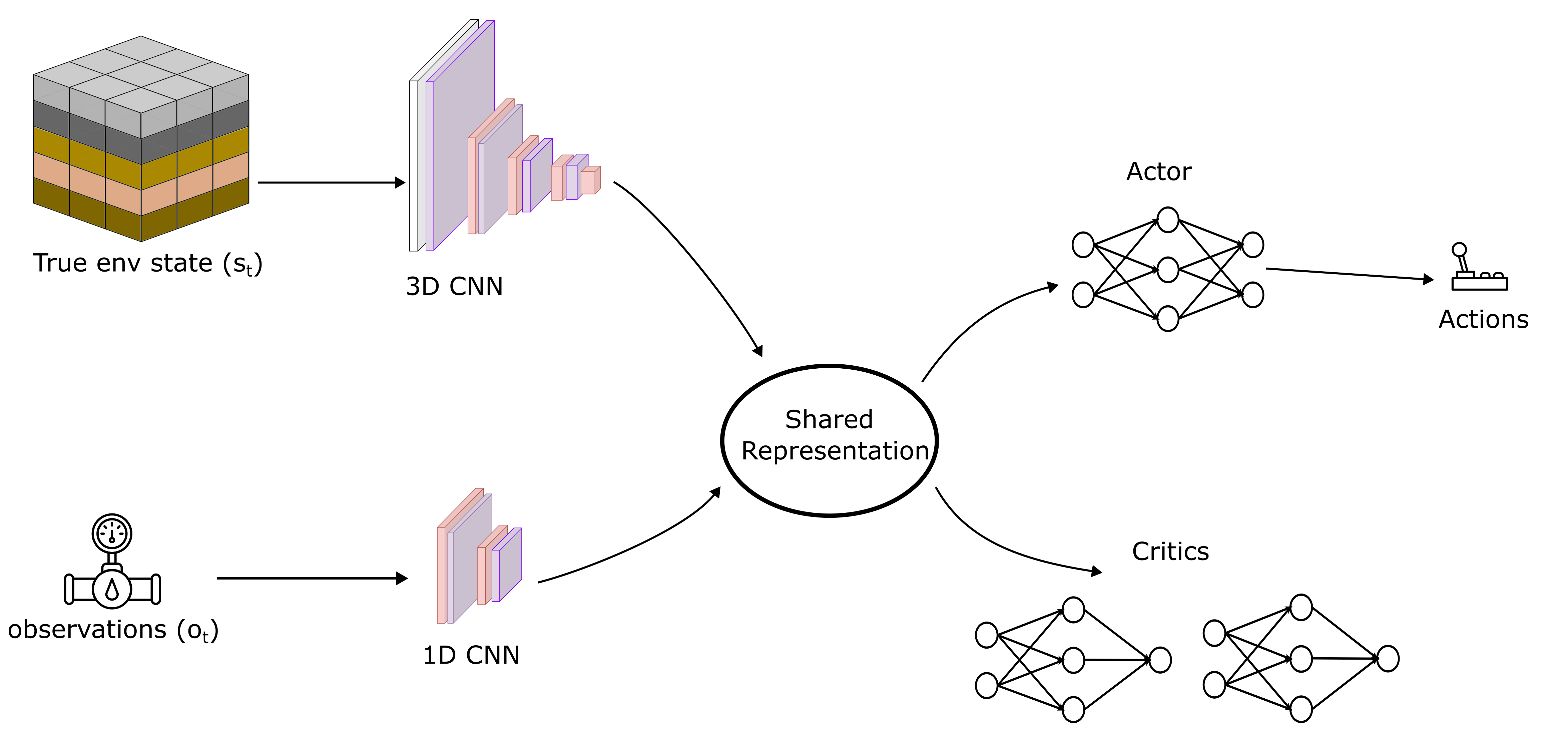}
	\caption{Baseline information regimes. The privileged-state benchmark uses dense simulator fields and well observations, whereas the well-only baseline removes the spatial branch and acts only from well-level observations.}
	\label{fig:mf_observation_regimes}
\end{figure}

The privileged-state benchmark uses the target-reservoir observation
\begin{equation}
	o_t^{\mathrm{priv}} =
	\left(
	x_t^{\mathrm{field}},\, y_t^{\mathrm{well}}
	\right),
\end{equation}
where $x_t^{\mathrm{field}}$ is the dense simulator-derived reservoir tensor and $y_t^{\mathrm{well}}$ is the most recent well-response summary. The spatial tensor is
\begin{equation}
	x_t^{\mathrm{field}} \in \mathbb{R}^{2 \times 4 \times 163 \times 120},
\end{equation}
with channels corresponding to normalized pressure and a saturation-related field. The well-observation tensor is
\begin{equation}
	y_t^{\mathrm{well}} \in \mathbb{R}^{9 \times 30}.
\end{equation}
It contains nine intra-interval samples over the latest two-year control interval. Each sample contains the following well measurements: gas injection rate, gas breakthrough rate, water production rate and well bottomhole pressures. This intra-interval sampling provides a secondary time level within each control step, but it does not by itself provide memory across previous control decisions.

Architecturally, the privileged actor and critics use two observation branches. The spatial branch applies a 3D convolutional encoder to $x_t^{\mathrm{field}}$, with four convolutional layers followed by global average pooling. The well branch applies a one-dimensional convolution over the nine intra-interval samples of $y_t^{\mathrm{well}}$. The resulting spatial and well features are concatenated and passed through two fully connected layers before the SAC actor and critic heads. For the critics, the action is concatenated with the encoded observation before value estimation. A schematic is shown in Figure~\ref{fig:mf_observation_regimes}.

\subsubsection{Well-only baseline}

The well-only baseline removes the dense simulator state and uses
\begin{equation}
	o_t^{\mathrm{well}} = y_t^{\mathrm{well}}.
\end{equation}
This controller is trained on prior digital-twin realizations and evaluated on the held-out target reservoir using only sparse well-level observations.

The actor and critics retain only the one-dimensional well-observation encoder and the final fully connected SAC heads. Because no spatial pressure/saturation tensor or inter-step memory is provided, this baseline isolates the difficulty of acting from the current well response alone.

\subsection{History-conditioned sequence model}

The well-only baseline uses only the most recent well-response tensor and is therefore limited in a partially observable setting. Similar current well measurements may correspond to different reservoir states depending on the preceding control sequence, pressure evolution, and displacement history. The history-conditioned variant addresses this limitation by conditioning the policy and critics on a finite history of public well observations.

The policy input is extended from
\begin{equation}
	o_t^{\text{well}} = y_t^{\text{well}}
\end{equation}
to
\begin{equation}
	o_t^{\text{hist}} = \left(y_t^{\text{well}},\, H_t\right),
\end{equation}
where
\begin{equation}
	H_t =
	\left[
	y_{t-L+1}^{\text{well}}, \ldots, y_t^{\text{well}}
	\right]
	\in \mathbb{R}^{L \times 9 \times 30}
\end{equation}
is a rolling history of the most recent $L$ decision-step observations. In this work, $L=20$, equal to the episode horizon, so the controller can access the full sequence of well responses accumulated during an episode. At reset, the history buffer is initialized with zeros and is then updated after each simulator step.

Each well-observation block $y_t^{\text{well}} \in \mathbb{R}^{9 \times 30}$ contains nine uniformly sampled intra-interval snapshots of the 30 well variables described previously. Rather than reprocessing the full raw history directly, each block is first mapped to a compact embedding by a shared one-dimensional convolutional encoder. In the implementation, the tensor is interpreted as 30 well-variable channels over nine intra-interval samples. A one-dimensional convolution with kernel size 3 and 64 output channels is followed by a ReLU activation and average pooling over the intra-interval dimension, giving
\begin{equation}
	e_t = f_{\mathrm{conv}}\!\left(y_t^{\text{well}}\right) \in \mathbb{R}^{64}.
\end{equation}
Applying the same encoder to every element of the history gives a sequence of embeddings
\begin{equation}
	E_t =
	\left[
	e_{t-L+1}, \ldots, e_t
	\right]
	\in \mathbb{R}^{L \times 64}.
\end{equation}

\begin{figure}[t]
	\centering
	\includegraphics[width=0.95\linewidth]{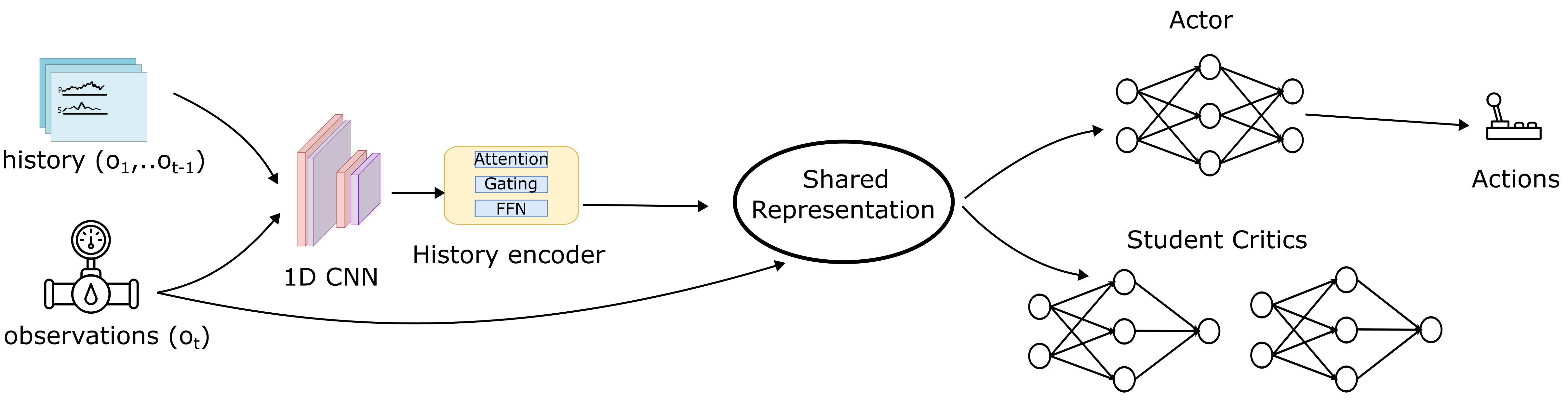}
	\caption{History-conditioned model. Current well observations and a rolling history of public well observations are encoded through a shared 1D convolutional front-end. The history embeddings are processed by a gated transformer-style sequence encoder, whose final output is fused with the current observation embedding and used by the SAC actor and twin critics.}
	\label{fig:history_model}
\end{figure}

Temporal dependencies across control steps are modeled using a gated transformer-style encoder applied to the explicitly stored history tensor. Importantly, this module is not used as a stateful recurrent policy: no hidden state is carried across environment steps or through the replay buffer. Instead, each replay-buffer sample already contains the finite history $H_t$, and the sequence encoder deterministically maps this history to a control feature. For an input embedding sequence $E_t$, the attention block computes multi-head self-attention over the decision-step history,
\begin{equation}
	A_t = \mathrm{MHA}\!\left(\mathrm{LN}(E_t)\right).
\end{equation}
The attention output is then combined with the incoming representation through a GRU-gated residual connection,
\begin{equation}
	\tilde{E}_t = \mathrm{GRUGate}(E_t, A_t),
\end{equation}
rather than by a standard additive residual connection. Here, the GRU cell is used only as a gating mechanism for the residual pathway, not as an external recurrent memory. The same gated residual structure is applied to the feedforward block,
\begin{equation}
	\bar{E}_t =
	\mathrm{GRUGate}
	\left(
	\tilde{E}_t,\,
	\mathrm{FFN}\!\left(\mathrm{LN}(\tilde{E}_t)\right)
	\right).
\end{equation}
The feedforward module uses a hidden dimension of 256 and maps back to the 64-dimensional embedding space. In the present implementation, one gated transformer-style block is used, with four attention heads of dimension 16. The final history feature is taken as the last output of the sequence encoder,
\begin{equation}
	h_t = \bar{E}_{t,L} \in \mathbb{R}^{64}.
\end{equation}

This design should therefore be interpreted as finite-history sequence encoding for off-policy SAC rather than as recurrent-state learning. The temporal information is supplied by the environment through $H_t$, while the gated transformer-style encoder learns how to summarize that finite observation history.

The current observation is also encoded separately using the same convolutional encoder, giving $e_t^{\mathrm{cur}} = f_{\mathrm{conv}}(y_t^{\text{well}})$. The actor representation is then formed by concatenating the current and history features,
\begin{equation}
	\phi_t^{\text{hist}} =
	\left[
	e_t^{\mathrm{cur}},\, h_t
	\right],
\end{equation}
and passing the resulting 128-dimensional vector through two fully connected layers of width 256. The actor outputs the mean and standard deviation of the Gaussian SAC policy. The twin critics use the same history-conditioned representation, concatenated with the action, before estimating the corresponding state-action values. Because the encoder uses only $(H_t, y_t^{\mathrm{well}})$, the policy preserves the deployable observation interface defined in Section~3.2. Figure~\ref{fig:history_model} summarizes the architecture.

\subsection{Curriculum masking of privileged state information}
\begin{figure}[t]
    \centering
    \includegraphics[width=0.95\linewidth]{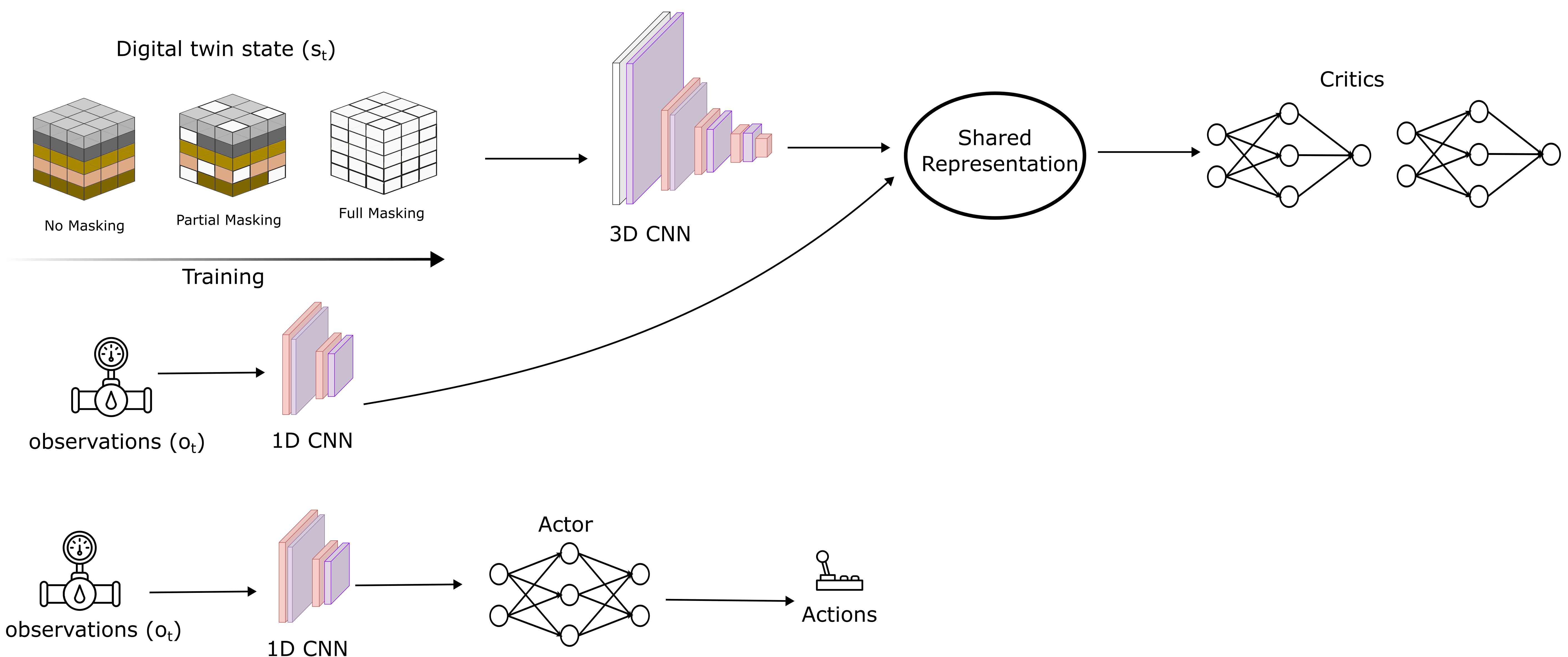}
    \caption{Masked-critic curriculum. During training, the critic receives progressively masked spatial simulator information, ranging from no masking to full masking, while the actor remains restricted to public well observations throughout. This yields an asymmetric actor-critic setup that gradually transitions from privileged to fully deployable training.}
    \label{fig:masking_curriculum}
\end{figure}

The masking-curriculum variant keeps the actor well-only but allows the critics to use a spatial simulator tensor whose information content is progressively reduced during training. This creates an asymmetric actor-critic diagnostic: the policy remains deployable, while the value estimators are tested under decreasing access to privileged state information. Figure~\ref{fig:masking_curriculum} summarizes the setup.

As in the previous variants, the environment returns a dense spatial reservoir tensor together with a well-observation tensor,
\begin{equation}
o_t = \left(x_t^{\text{field}},\, y_t^{\text{well}}\right),
\end{equation}
with
\begin{equation}
x_t^{\text{field}} \in \mathbb{R}^{2 \times 4 \times 163 \times 120},
\qquad
y_t^{\text{well}} \in \mathbb{R}^{9 \times 30}.
\end{equation}
Before the spatial tensor is used by the learning algorithm, however, a random masking operator is applied:
\begin{equation}
\tilde{x}_t^{\text{field}} = M_p\!\left(x_t^{\text{field}}\right),
\end{equation}
where $M_p(\cdot)$ denotes stochastic masking with ratio $p \in [0,1]$. Specifically, a fraction $p$ of the nonzero entries of the spatial tensor is selected uniformly at random and set to zero. Hence, $p=0$ corresponds to the unmodified spatial state, whereas $p=1$ corresponds to full masking, i.e., a zero spatial tensor. The well observations remain unchanged throughout.

A key design choice in this experiment is that the actor and critics do not receive the same inputs. The actor is intentionally restricted to well observations only and is therefore parameterized as
\begin{equation}
\pi_\theta(a_t \mid y_t^{\text{well}}),
\end{equation}
using the same one-dimensional convolutional encoder employed in the well-only baseline. In contrast, both critics receive the masked spatial tensor together with the well observations and action,
\begin{equation}
Q_{\phi_i}\!\left(\tilde{x}_t^{\text{field}}, y_t^{\text{well}}, a_t\right), \qquad i \in \{1,2\},
\end{equation}
and process the spatial component through a 3D convolutional encoder before combining it with well features and the action. This yields an asymmetric actor-critic setting: the policy remains deployable from well data alone, while the critics are allowed to exploit additional spatial information during training, albeit in progressively degraded form.

The masking ratio is not fixed, but instead follows a curriculum over training epochs,
\begin{equation}
p \in \{0.00,\;0.25,\;0.50,\;0.75,\;1.00\}.
\end{equation}
Training begins with $p=0$, so that the critics observe the full privileged spatial state. The masking level is then increased in stages until the critics are trained with fully masked spatial input. Importantly, the replay buffer is reset whenever the masking ratio changes. This prevents transitions collected under one observability regime from being mixed with transitions collected under another and thereby reduces distribution mismatch during off-policy updates. During evaluation, the masking ratio is fixed at
\begin{equation}
p_{\text{test}} = 1.0,
\end{equation}
so that test-time performance reflects the fully deployable regime in which no spatial simulator field is available.

We use this curriculum as a diagnostic stress test of critic-side information dependence. Since the actor is always restricted to $y_t^{\mathrm{well}}$, the deployment interface does not change across masking levels. The experiment therefore measures whether progressively removing privileged critic information destabilizes learning or whether well-based features are sufficient to support the deployable actor.

\subsection{Asymmetric teacher-student sequence model}

\begin{figure}[t]
	\centering
	\includegraphics[width=\linewidth]{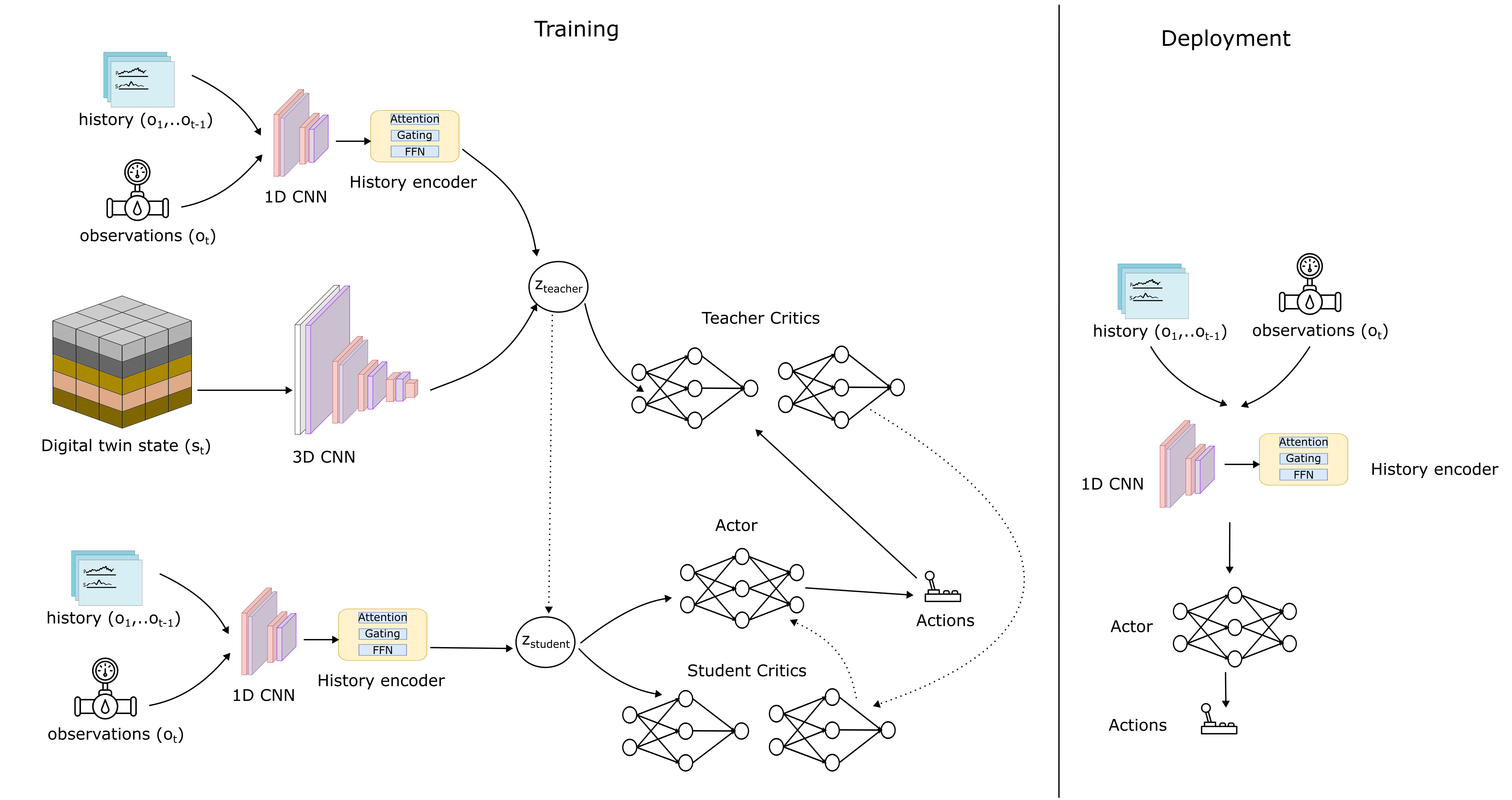}
	\caption{Asymmetric teacher-student model. During training, privileged teacher critics use dense spatial simulator fields together with public well-history information. Student critics and the actor use only public current and historical well observations. The actor is updated through the student critics, while an auxiliary distillation loss encourages the public history encoder to capture information contained in the privileged spatial branch. At deployment, only the public history encoder and actor are used.}
	\label{fig:asymmetric_pomdp_model}
\end{figure}

The final model-free variant combines history-based public control, asymmetric critic learning, and auxiliary latent distillation from privileged simulator states. Unlike the masking curriculum, privileged information is not gradually removed; it is confined to teacher critics, while the actor and student critics use only public well-history observations. Figure~\ref{fig:asymmetric_pomdp_model} summarizes the setup.

At each decision step, the simulator provides
\begin{equation}
	o_t = \left(x_t^{\mathrm{field}},\, y_t^{\mathrm{well}},\, H_t\right),
\end{equation}
where $x_t^{\mathrm{field}}$ is the dense spatial simulator state, $y_t^{\mathrm{well}}$ is the current well-observation tensor, and $H_t$ is the rolling history of public well observations. The public observation available to the deployable policy is
\begin{equation}
	o_t^{\mathrm{pub}} =
	\left(H_t,\, y_t^{\mathrm{well}}\right),
\end{equation}
whereas the privileged training observation used by the teacher critics is
\begin{equation}
	o_t^{\mathrm{teach}} =
	\left(x_t^{\mathrm{field}},\, H_t,\, y_t^{\mathrm{well}}\right).
\end{equation}

The actor uses the same finite-history idea introduced in the previous subsection, but with a larger latent dimension. A shared one-dimensional convolution maps each $9 \times 30$ well-observation block to a $d$-dimensional embedding, with $d=128$ in the present implementation. The sequence of historical embeddings is processed by a gated transformer-style encoder. The final history embedding is then combined with the current-observation embedding by addition,
\begin{equation}
	\phi_t^{\mathrm{pub}}
	=
	f_{\mathrm{hist}}\!\left(H_t, y_t^{\mathrm{well}}\right)
	\in \mathbb{R}^{128}.
\end{equation}
This public feature is passed to a policy head that outputs the mean and standard deviation of the Gaussian SAC policy. The two teacher critics receive the privileged observation and action. Each teacher critic has a spatial branch and a public-history branch. The spatial branch maps the dense simulator tensor to a privileged latent vector,
\begin{equation}
	z_{t,i}^{\mathrm{teach}}
	=
	f_{3D,i}\!\left(x_t^{\mathrm{field}}\right),
\end{equation}
while the public-history branch maps the well-history observation to
\begin{equation}
	h_{t,i}^{\mathrm{teach}}
	=
	f_{\mathrm{hist},i}^{Q}
	\left(H_t, y_t^{\mathrm{well}}\right).
\end{equation}
The teacher critic value is then computed as
\begin{equation}
	Q_i^{\mathrm{teach}}
	\left(o_t^{\mathrm{teach}}, a_t\right)
	=
	q_i^{\mathrm{teach}}
	\left(
	z_{t,i}^{\mathrm{teach}},
	h_{t,i}^{\mathrm{teach}},
	a_t
	\right),
	\qquad i \in \{1,2\}.
\end{equation}
Thus, the teacher critics can exploit dense pressure/saturation information when constructing value estimates during training.

In parallel, two student critics are trained using only public inputs,
\begin{equation}
	Q_i^{\mathrm{stud}}
	\left(o_t^{\mathrm{pub}}, a_t\right)
	=
	q_i^{\mathrm{stud}}
	\left(
	f_{\mathrm{hist},i}^{\mathrm{stud}}
	\left(H_t, y_t^{\mathrm{well}}\right),
	a_t
	\right),
	\qquad i \in \{1,2\}.
\end{equation}
Only the teacher critics use $x_t^{\mathrm{field}}$; the actor and student critics share the public observation interface $o_t^{\mathrm{pub}}$. This is important because the actor update is computed through the student critics, not through the privileged teacher critics. As a result, privileged information can improve the training target, but the policy gradient remains tied to value estimates that are realizable from deployment-time observations.

Training proceeds as follows within each SAC update. First, the teacher critics are updated using the standard SAC critic loss on privileged observations. Then, a teacher TD target is constructed using the target teacher critics and the next action sampled from the public actor:
\begin{equation}
	y_t^{\mathrm{teach}}
	=
	r_t
	+
	(1-d_t)\gamma
	\left[
	\min_{i \in \{1,2\}}
	Q_{i,\mathrm{target}}^{\mathrm{teach}}
	\left(
	o_{t+1}^{\mathrm{teach}}, a_{t+1}
	\right)
	-
	\alpha
	\log
	\pi
	\left(
	a_{t+1}
	\mid
	o_{t+1}^{\mathrm{pub}}
	\right)
	\right].
\end{equation}
The student critics are then trained to regress to this privileged target:
\begin{equation}
	\mathcal{L}_{Q}^{\mathrm{stud}}
	=
	\sum_{i=1}^{2}
	\left\|
	Q_i^{\mathrm{stud}}
	\left(
	o_t^{\mathrm{pub}}, a_t
	\right)
	-
	y_t^{\mathrm{teach}}
	\right\|_2^2.
\end{equation}
Finally, the actor is updated using the student critics:
\begin{equation}
	\mathcal{L}_{\pi}
	=
	\mathbb{E}
	\left[
	\alpha
	\log
	\pi
	\left(
	a_t \mid o_t^{\mathrm{pub}}
	\right)
	-
	\min_{i \in \{1,2\}}
	Q_i^{\mathrm{stud}}
	\left(
	o_t^{\mathrm{pub}}, a_t
	\right)
	\right].
\end{equation}
This update structure separates the source of the training target from the source of the actor gradient: teacher critics provide privileged TD targets, student critics approximate those targets from public observations, and the actor learns through the public student branch.

An auxiliary distillation loss is added to encourage the public history representation to contain information correlated with the privileged spatial latent. The student distillation pathway shares the actor history encoder and applies a distillation head to produce
\begin{equation}
	z_t^{\mathrm{stud}}
	=
	g_{\mathrm{dist}}
	\left(
	f_{\mathrm{hist}}
	\left(H_t, y_t^{\mathrm{well}}\right)
	\right).
\end{equation}
The target latent is obtained from the spatial encoder of the first teacher critic,
\begin{equation}
	z_t^{\mathrm{teach}}
	=
	f_{3D,1}
	\left(x_t^{\mathrm{field}}\right),
\end{equation}
with gradients stopped through the teacher branch. In the present implementation, projected and normalized student and teacher latents are matched using an InfoNCE-style contrastive loss,
\begin{equation}
	\mathcal{L}_{\mathrm{NCE}}
	=
	-\frac{1}{B}
	\sum_{b=1}^{B}
	\log
	\frac{
		\exp
		\left(
		\langle p_s(z_{b}^{\mathrm{stud}}), p_t(z_{b}^{\mathrm{teach}}) \rangle / \tau
		\right)
	}{
		\sum_{j=1}^{B}
		\exp
		\left(
		\langle p_s(z_{b}^{\mathrm{stud}}), p_t(z_{j}^{\mathrm{teach}}) \rangle / \tau
		\right)
	},
\end{equation}
where $p_s$ is trainable, $p_t$ is kept frozen, and $\tau$ is the contrastive temperature. A scalar value-alignment head is also trained from the student latent to match the detached student-side value estimate. The auxiliary distillation objective is therefore
\begin{equation}
	\mathcal{L}_{\mathrm{dist}}
	=
	\mathcal{L}_{\mathrm{NCE}}
	+
	\beta
	\mathcal{L}_{\mathrm{val}}.
\end{equation}
The total actor-side loss used when distillation is active is
\begin{equation}
	\mathcal{L}_{\mathrm{actor,total}}
	=
	\mathcal{L}_{\pi}
	+
	w_{\mathrm{dist}}(e)
	\mathcal{L}_{\mathrm{dist}},
\end{equation}
where $w_{\mathrm{dist}}(e)$ is linearly annealed over the final part of training. In the reported configuration, distillation is ramped from epoch 22 to epoch 30 with final weight 0.1. At test time, only the public history encoder and policy head are used; teacher critics, student critics, spatial encoders, and distillation heads are discarded.

This variant therefore combines the public history interface of Section~3.4 with asymmetric privileged value supervision, making it the most structured deployable model-free controller considered here.

\section{Latent model-based adaptation}

Section~3 considered deployable model-free controllers trained directly with high-fidelity simulator interaction. We now test whether a latent model-based controller can retain similar real-simulator performance while reducing the amount of scenario-specific simulator interaction needed for adaptation. The fixed pipeline maps public well-history observations to a latent control state, trains a latent world model on simulator-derived transitions, and updates a latent actor-critic pair through Dreamer-style imagined rollouts.

\begin{figure}[t]
    \centering
    \includegraphics[width=\linewidth]{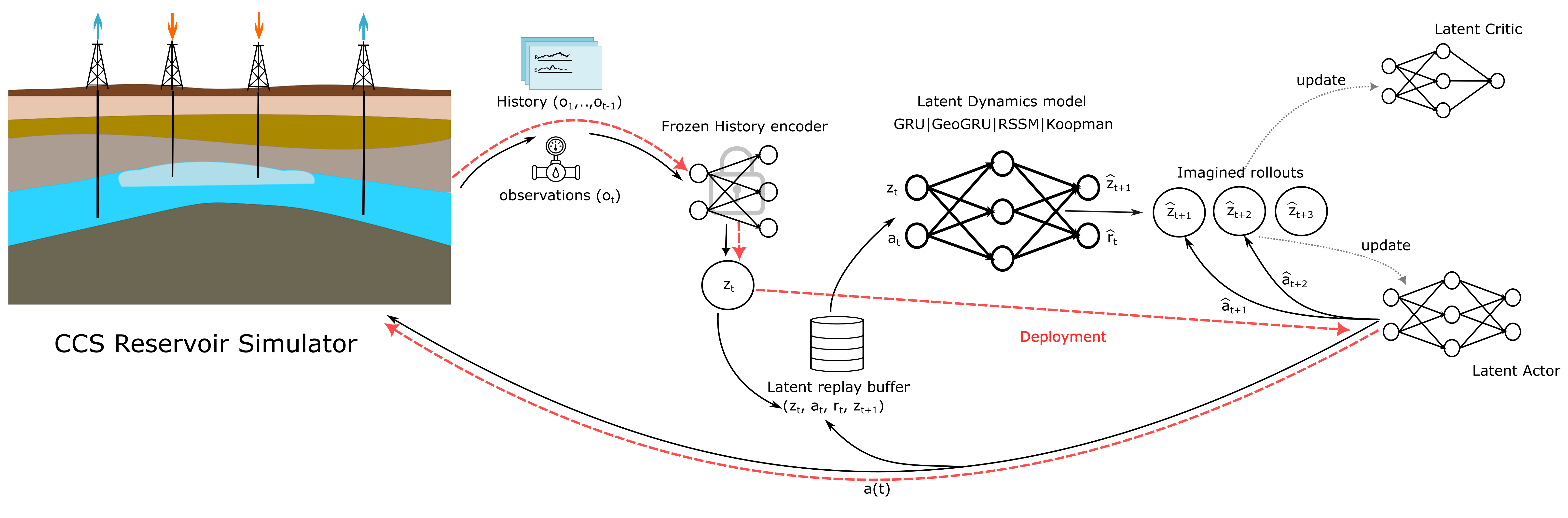}
    \caption{Model-based pipeline used in this work. Public observations are mapped to a deployable latent state through a frozen history encoder, latent transitions are stored in a replay buffer, and a latent dynamics model is trained to support Dreamer-style imagined rollouts for the latent actor and critic. The same latent actor is also deployed back in the real CCS simulator.}
    \label{fig:mb_pipeline}
\end{figure}

Within this fixed pipeline, we consider four latent world-model backbones, described in the next subsection: a probabilistic GRU, a geometrically regularized GRU, an RSSM, and a Koopman-inspired model. These backbones are evaluated under the same overall training and control framework across all scenarios. A schematic overview of the full model-based pipeline is shown in Figure~\ref{fig:mb_pipeline}.

\subsection{Latent world-model backbones}
\label{sec:world_models}

The world model predicts the next deployable latent state and the immediate
reward from the current latent state and control action. Let
$z_t \in \mathbb{R}^{d_z}$ denote the latent representation produced by the
frozen public history encoder, $a_t \in \mathbb{R}^{d_a}$ the normalized
control action, and $r_t$ the reward. The training data consist of latent
transition tuples
\begin{equation}
	\mathcal{D}
	=
	\left\{
	(z_t, a_t, r_t, z_{t+1})
	\right\},
\end{equation}
extracted from simulator interaction. Each backbone predicts
\begin{equation}
	(z_t, a_t) \longmapsto (\hat{z}_{t+1}, \hat{r}_t),
\end{equation}
possibly conditioned on an internal recurrent state. The learned transition
model is subsequently used to generate imagined trajectories for latent
actor--critic updates.

We compare four world-model backbones with distinct assumptions about the
latent dynamics: a probabilistic GRU, a geometrically regularized GRU, an
RSSM-inspired model, and a controlled linear-residual
Koopman-inspired model.

\paragraph{Probabilistic GRU.}
The first backbone uses a gated recurrent unit (GRU) to summarize recent
latent-action history. Given recurrent state $h_t$, the model computes
\begin{equation}
	\begin{aligned}
		x_t &= \operatorname{ReLU}\!\left(
		W_x [z_t ; a_t] + b_x
		\right), \\
		h_{t+1} &= \operatorname{GRU}(x_t,h_t), \\
		[\mu_t ; \tilde{\sigma}_t]
		&= W_z h_{t+1} + b_z, \\
		\sigma_t &= \operatorname{softplus}(\tilde{\sigma}_t) + 10^{-3}, \\
		\hat{r}_t &= r_\theta(h_{t+1}),
	\end{aligned}
	\label{eq:gru_world_model}
\end{equation}
which defines a diagonal Gaussian transition model,
\begin{equation}
	p_\theta(z_{t+1} \mid z_t,a_t,h_t)
	=
	\mathcal{N}
	\left(
	\mu_t,
	\operatorname{diag}(\sigma_t^2)
	\right).
\end{equation}
For a sequence of length $L$, the training objective is
\begin{equation}
	\mathcal{L}_{\mathrm{GRU}}
	=
	\frac{1}{L}
	\sum_{t=0}^{L-1}
	\left[
	\ell_{\mathrm{NLL}}
	\left(
	z_{t+1};
	\mu_t,\sigma_t^2
	\right)
	+
	\left\|
	\hat{r}_t-r_t
	\right\|_2^2
	\right],
	\label{eq:gru_loss}
\end{equation}
where $\ell_{\mathrm{NLL}}$ is the Gaussian negative log-likelihood. During
imagination, next latents are sampled as
\begin{equation}
	\hat{z}_{t+1}
	=
	\mu_t + \sigma_t \odot \epsilon_t,
	\qquad
	\epsilon_t \sim \mathcal{N}(0,I).
\end{equation}

\paragraph{Geometry-aware GRU.}
The second backbone uses the same probabilistic GRU parameterization, but
adds a decoder-induced local geometric penalty. Let
\begin{equation}
	D_{\mathrm{eff}}:
	\mathbb{R}^{d_z}
	\rightarrow
	\mathbb{R}^{M}
\end{equation}
denote a frozen decoder from latent space to the reconstructed spatial field.
For post-finetuning latents, $D_{\mathrm{eff}}$ is the spatial decoder itself.
For pre-finetuning latents, it is composed with the learned pre-to-post latent
mapping before decoding.

For a predicted mean $\mu_t$, define the latent prediction error
\begin{equation}
	e_t = \mu_t-z_{t+1}.
\end{equation}
The geometric term measures the local decoded consequence of this error,
\begin{equation}
	\ell_{\mathrm{geom},t}
	=
	\frac{1}{M}
	\left\|
	J_{D_{\mathrm{eff}}}
	\left(
	\operatorname{sg}[\mu_t]
	\right)
	e_t
	\right\|_2^2,
	\label{eq:geometric_loss}
\end{equation}
where $J_{D_{\mathrm{eff}}}$ is the decoder Jacobian and
$\operatorname{sg}[\cdot]$ denotes stop-gradient. Equivalently, this term
weights latent errors by the local pullback metric
$J_{D_{\mathrm{eff}}}^{\top}J_{D_{\mathrm{eff}}}$. Thus, latent directions
that produce larger changes in the decoded spatial field receive a larger
penalty.

The geometry-aware objective is
\begin{equation}
	\mathcal{L}_{\mathrm{GeoGRU}}
	=
	\ell_{\mathrm{NLL}}
	+
	\lambda_{\mathrm{geom}}
	\ell_{\mathrm{geom}}
	+
	\left\|
	\hat{r}_t-r_t
	\right\|_2^2,
	\label{eq:geogru_loss}
\end{equation}
with $\lambda_{\mathrm{geom}}=0.05$. To limit memory cost, the Jacobian-vector
product in Eq.~\eqref{eq:geometric_loss} is evaluated on at most 32 randomly
selected transitions per mini-batch.

In the present implementation, this loss is evaluated transition-wise:
sampled sequences are flattened and the GRU hidden state is initialized to zero
for each transition. Therefore, although the model uses a GRU transition
parameterization, the geometry-aware objective does not exploit recurrent
history during world-model fitting.

\paragraph{RSSM-inspired model.}
The third backbone is a recurrent state-space model with a deterministic
recurrent state $h_t$ and a stochastic latent state $s_t$. Its deterministic
dynamics are
\begin{equation}
	h_{t+1}
	=
	\operatorname{GRU}
	\left(
	\phi_\theta([z_t;a_t]),
	h_t
	\right),
\end{equation}
where $\phi_\theta$ is a learned input projection. The model defines a
diagonal-Gaussian prior
\begin{equation}
	p_\theta(s_{t+1}\mid h_{t+1})
	=
	\mathcal{N}
	\left(
	\mu^p_t,
	\operatorname{diag}((\sigma^p_t)^2)
	\right),
\end{equation}
and, during training, an inference posterior conditioned on the observed next
latent,
\begin{equation}
	q_\theta(s_{t+1}\mid h_{t+1},z_{t+1})
	=
	\mathcal{N}
	\left(
	\mu^q_t,
	\operatorname{diag}((\sigma^q_t)^2)
	\right).
\end{equation}
A stochastic state is drawn through reparameterization,
\begin{equation}
	s_{t+1}
	=
	\mu^q_t+\sigma^q_t\odot\epsilon_t,
	\qquad
	\epsilon_t\sim\mathcal{N}(0,I),
\end{equation}
and is used together with the deterministic state to predict the next
observed latent and reward:
\begin{equation}
	\hat{z}_{t+1}
	=
	g_\theta([h_{t+1};s_{t+1}]),
	\qquad
	\hat{r}_t
	=
	r_\theta([h_{t+1};s_{t+1}]).
\end{equation}
The sequence objective is
\begin{equation}
	\begin{aligned}
		\mathcal{L}_{\mathrm{RSSM}}
		=
		\frac{1}{L}
		\sum_{t=0}^{L-1}
		\Bigg[
		&
		\left\|
		\hat{z}_{t+1}-z_{t+1}
		\right\|_2^2
		+
		\left\|
		\hat{r}_t-r_t
		\right\|_2^2
		\\&
		+
		\beta
		D_{\mathrm{KL}}
		\left(
		q_\theta(s_{t+1}\mid h_{t+1},z_{t+1})
		\,\Vert\,
		p_\theta(s_{t+1}\mid h_{t+1})
		\right)
		\Bigg],
	\end{aligned}
	\label{eq:rssm_loss}
\end{equation}
with $\beta=1$ in the reported implementation. During imagined rollouts, the
stochastic state is set to the prior mean,
\begin{equation}
	s_{t+1}=\mu^p_t,
\end{equation}
so RSSM rollouts are deterministic conditional on the selected ensemble
member.

\paragraph{Koopman-inspired linear-residual model.}
The fourth backbone assumes that the latent dynamics can be represented mainly
by a controlled linear transition, augmented by a scaled nonlinear residual:
\begin{equation}
	\hat{z}_{t+1}
	=
	A z_t
	+
	B a_t
	+
	\alpha\epsilon_\theta([z_t;a_t]),
	\label{eq:koopman_transition}
\end{equation}
where $\epsilon_\theta$ is a multilayer perceptron and $\alpha=0.1$. The
reward model is
\begin{equation}
	\hat{r}_t
	=
	r_\theta([\hat{z}_{t+1};a_t]).
\end{equation}
This model is trained on one-step transitions using
\begin{equation}
	\mathcal{L}_{\mathrm{K}}
	=
	\left\|
	\hat{z}_{t+1}-z_{t+1}
	\right\|_2^2
	+
	\left\|
	\hat{r}_t-r_t
	\right\|_2^2
	+
	\lambda_{\mathrm{K}}
	\left(
	\|A\|_F^2+\|B\|_F^2
	\right),
	\label{eq:koopman_loss}
\end{equation}
where $\lambda_{\mathrm{K}}=10^{-4}$. The linear component and matrix
regularization provide a stabilizing bias, while the residual preserves
capacity for nonlinear latent dynamics. We therefore refer to this model as
\emph{Koopman-inspired}, rather than as an exact Koopman operator model.

\paragraph{Training and imagined rollouts.}
The standard GRU and RSSM are trained on overlapping episode-respecting latent
windows of length $L=10$, with $h_0=0$ at the beginning of each sampled
window. The Koopman-inspired model is trained on one-step transitions. Unless
otherwise stated, we use an ensemble of $K=10$ independently initialized
models of the same backbone. The ensemble objective is the average member
loss,
\begin{equation}
	\mathcal{L}_{\mathrm{ens}}
	=
	\frac{1}{K}
	\sum_{k=1}^{K}
	\mathcal{L}(\theta_k).
\end{equation}
During imagined rollouts, one ensemble member is selected uniformly at random
for each model transition. The probabilistic GRU additionally samples from its
Gaussian transition distribution, whereas the RSSM and Koopman-inspired
backbones use deterministic transitions conditional on the selected member.

The default hidden dimensions are 512 for the GRU and RSSM deterministic
states, 32 for the RSSM stochastic state, and 256 for the Koopman residual
network. All backbones are optimized offline with Adam before being used in
the same latent Dreamer-style control loop.

\subsection{Scenario 0: nominal model-based retention of model-free performance}

Scenario~0 is the nominal retention test. No abnormality is introduced, the control interface is unchanged, and the reward is the same as in the model-free experiments. The purpose is to determine whether the fixed latent model-based pipeline of Figure~\ref{fig:mb_pipeline} can preserve the real-simulator performance of the deployable history-based model-free reference. We evaluate two public latent sources: the student history encoder obtained after teacher-student training, and the same encoder after post-hoc distillation toward the privileged spatial encoder. These are referred to as pre-finetuning and post-finetuning student latents, respectively.

The world model is pretrained offline on public latent transition tuples
\begin{equation}
	\left(z_t, a_t, r_t, z_{t+1}\right),
\end{equation}
extracted from prior simulator interaction. Recurrent backbones are trained on fixed-length latent sequences, whereas the Koopman-inspired model is trained on one-step transitions. After offline pretraining, the world model is inserted into a latent Dyna-style training loop with imagined rollouts. At each epoch, the latent actor collects a limited amount of fresh real simulator experience from training digital twins, encoded through the frozen public history encoder. These real latent transitions are used to continue updating the world model. The actor and critic are then trained using imagined trajectories generated by the latent dynamics model from sampled latent start states.

Evaluation is performed by deploying the learned latent actor back into the high-fidelity OPM/Flow simulator through the same public encoder. Real transitions used during model-based training are collected only from training digital-twin realizations, while performance is evaluated on the held-out target deck. Thus, Scenario~0 preserves the same train/test logic as the model-free study: prior realizations are used for learning, and the target environment is used for evaluation. Scenario~0 is therefore a viability benchmark: the world model need not be an exact surrogate of the full simulator, but it must be accurate enough in the public student latent space to support imagined rollouts that preserve real-simulator control performance.

\subsection{Abnormal-scenario retuning protocol}

Scenarios~1-3 evaluate adaptation after the nominal setting changes. In each abnormal scenario, the comparison is between two retuning strategies under the same scenario-specific real-simulator interaction budget. The first is direct model-free retuning in the abnormal environment. The second is latent model-based retuning, which reuses the nominal latent structure and updates the controller through a combination of limited real abnormal transitions and imagined rollouts. The abnormal scenarios differ only in the source of mismatch: a known loss of injector control authority, leakage-induced dynamics and reward shift, or changed reservoir connectivity.

For Scenarios~2 and~3, which involve abnormal physics rather than a known action clamp, we use a common residual-adapter protocol. A small abnormal-transition dataset is collected from the modified environment and encoded with the frozen public history encoder, giving tuples $(z_t,a_t,r_t,z_{t+1})$. The nominal Scenario~0 world model is frozen and used as prior structure. A residual adapter then corrects the nominal next-latent and reward predictions,
\begin{equation}
	\hat{z}_{t+1}^{\mathrm{corr}}
	=
	\hat{z}_{t+1}^{\mathrm{nom}}
	+
	\Delta z_{\psi},
	\qquad
	\hat{r}_{t}^{\mathrm{corr}}
	=
	\hat{r}_{t}^{\mathrm{nom}}
	+
	\Delta r_{\psi}.
\end{equation}
The corrected model is inserted into the latent Dyna-style retuning loop, where the actor is warm-started from the nominal deployable controller and updated using corrected imagined rollouts together with the limited abnormal real-transition budget.

\subsection{Scenario 1: known loss of one injector control channel}

Scenario~1 introduces an operational abnormality in which one injector control channel is no longer available for optimization. The reservoir physics and reward definition remain the same as in the nominal case, but the controller loses authority over injector $I_3$. This scenario therefore tests adaptation to a known degradation of the control interface, rather than adaptation to an unknown geological or physical mismatch.

The original 11-dimensional action parameterization is retained for compatibility with the nominal policy and latent world models. The failed control channel is handled by imposing a deterministic action mask on the last action component, corresponding to injector $I_3$. In normalized action space, this component is fixed to
\begin{equation}
	a_{t,10} = -1,
\end{equation}
which maps to the minimum admissible physical injection rate after unnormalization. The same convention is enforced in the simulator deck construction, where the injection rate of $I_3$ is overwritten by its lower bound. Thus, the policy may still output an 11-dimensional action, but the failed coordinate is clamped before the action is used by either the real simulator or the latent dynamics model.

For model-based retuning, the same action mask is applied during real abnormal data collection, latent world-model updates, imagined rollouts, and evaluation. This keeps the real and imagined failed-control regimes consistent. The scenario therefore asks whether latent model-based retuning can make better use of limited abnormal-regime simulator interaction than direct model-free retuning when the actuator loss is known and can be imposed explicitly.

\begin{figure}[t]
	\centering
	\includegraphics[width=\linewidth]{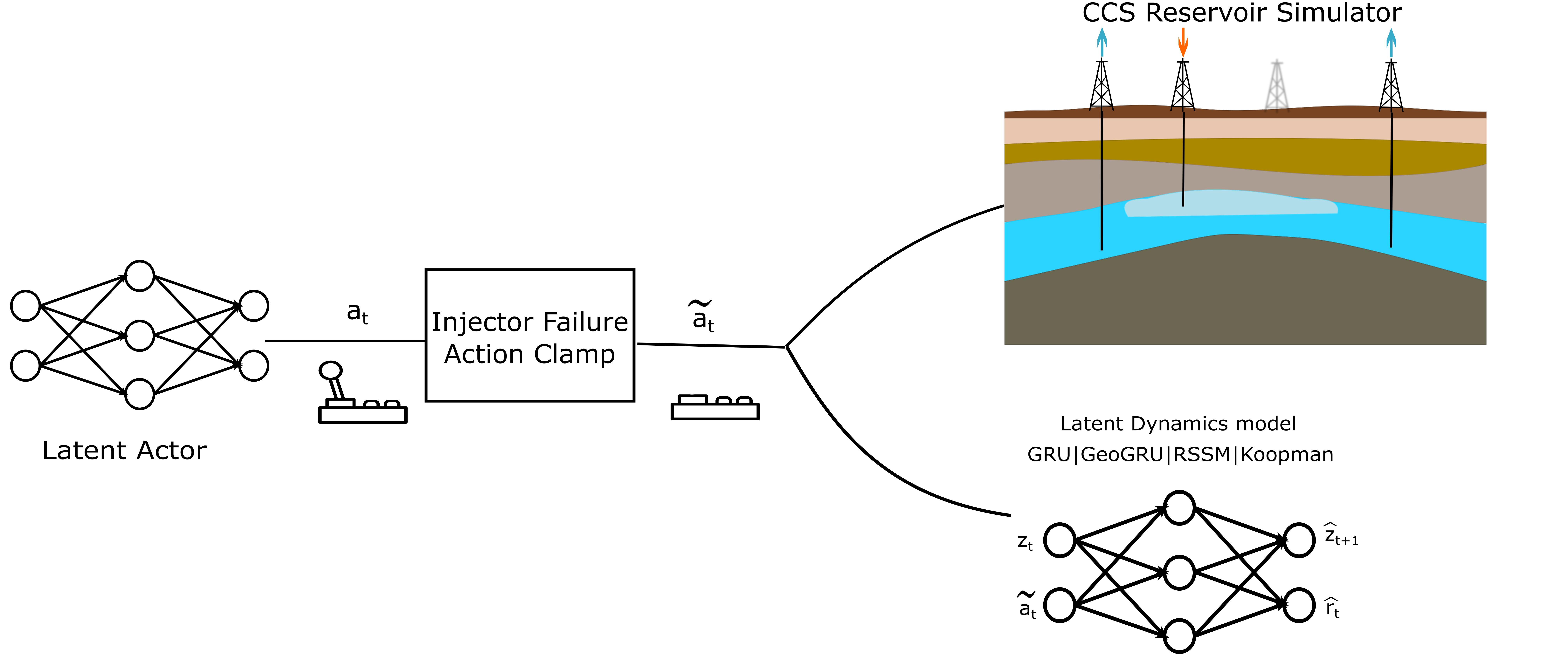}
	\caption{Scenario~1 methodology. The actor outputs an 11-dimensional nominal action, after which the failed-injector mask fixes the $I_3$ control component to its minimum admissible value. The same masked action is used in real simulator interaction, latent world-model updates, imagined rollouts, and evaluation, keeping the real and imagined failed-control regimes consistent.}
	\label{fig:scenario1_method}
\end{figure}

\subsection{Scenario 2: leakage-induced dynamics and reward shift}

Scenario~2 considers an abnormal-physics setting in which the nominal reservoir assumptions no longer describe the target environment. The action space and deployable observation interface remain unchanged, but the simulator is modified to represent a leakage pathway. As a result, the controller faces both a transition-dynamics shift and a reward shift. The public observations are still given by the current well responses and their rolling history, but the reward now includes an additional leakage penalty based on gas saturation in a designated leakage region. Once the leakage indicator exceeds a prescribed threshold, the reward is reduced accordingly.

Unlike Scenario~1, the abnormality is not a known action clamp that can be imposed exactly in both real and imagined interaction. Instead, the nominal latent dynamics model has become partially incorrect: the same public latent state and action may now lead to different latent transitions and rewards because leakage changes the underlying simulator response and the operational objective.

Scenario~2 applies the residual-adapter protocol of Section~4.3 to a leakage setting. Here the adapter must correct both latent transition dynamics and reward predictions because leakage changes the simulator response and introduces an additional penalty term.

\begin{figure}[t]
	\centering
	\includegraphics[width=\linewidth]{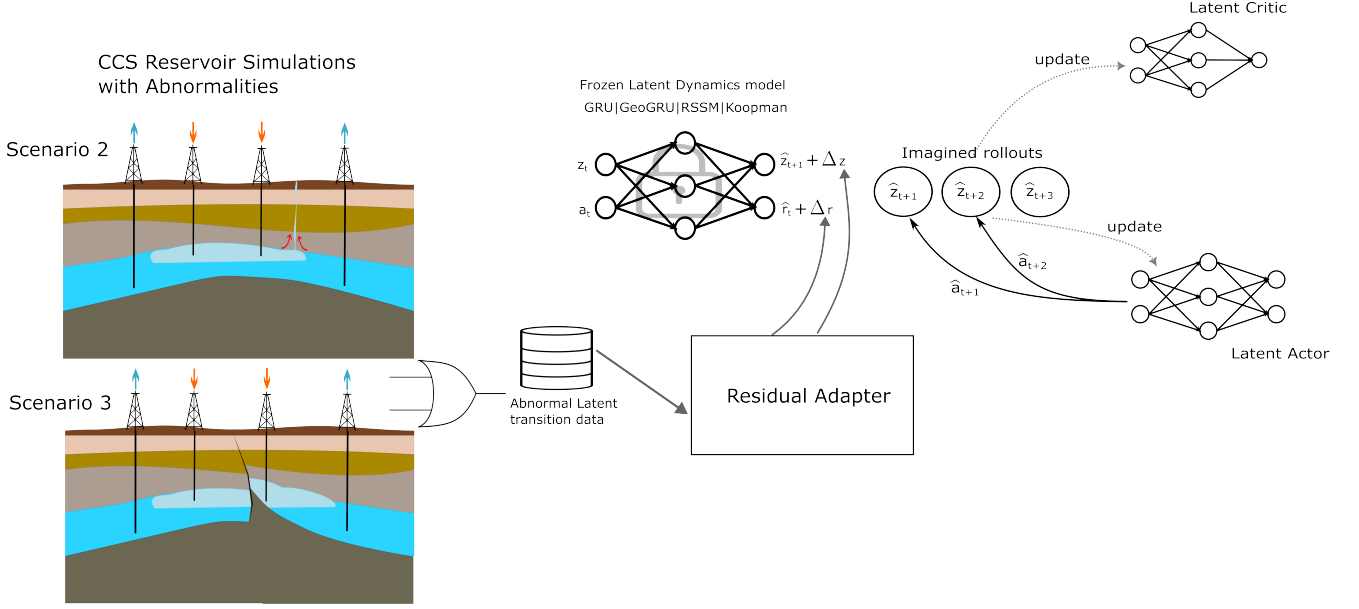}
	\caption{Residual world-model adaptation for Scenarios~2 and~3. Abnormal latent transitions are encoded using the frozen public history encoder. A residual adapter corrects nominal next-latent and reward predictions, and the corrected model is used for imagined rollouts during latent actor-critic retuning.}
	\label{fig:scenario23_method}
\end{figure}

Scenario~2 therefore tests residual model-based adaptation under a combined dynamics-and-reward shift.

\subsection{Scenario 3: compartmentalized reservoir with changed connectivity}

Scenario~3 considers a second abnormal-physics setting, in which the nominal reservoir model is assumed to be overly connected. Under the revised interpretation, the reservoir is more compartmentalized: communication between some injector-producer pairs is reduced because transmissibility across compartments is strongly restricted. The action space, reward definition, and deployable observation interface remain unchanged, but the same control sequence now produces different pressure propagation and flow responses. Thus, unlike Scenario~2, the main abnormality is a transition-dynamics shift rather than a combined dynamics-and-reward shift.

Scenario~3 uses the same residual-adapter protocol defined in Section~4.3. Since the reward definition is unchanged, the adapter primarily corrects latent transition dynamics induced by the revised connectivity pattern; reward correction is retained only for consistency with the general adapter formulation. It therefore tests residual model-based adaptation under a structural connectivity shift with an unchanged reward.

\section{Results}

We present the results in two stages. First, we compare the model-free controllers under different observability regimes in order to quantify the value of temporal context and privileged training information. Second, we examine whether the model-based pipeline can (i) retain the performance of the deployable model-free reference under nominal conditions and (ii) adapt more effectively than direct model-free retuning under abnormal operating conditions. For the abnormal-scenario figures, epoch 0 corresponds to the nominal controller evaluated in the abnormal environment before any scenario-specific updates.

Because each training run requires repeated high-fidelity OPM/Flow simulations, the reported curves and tables should be interpreted as simulator-budget-controlled comparisons rather than full random-seed sensitivity studies. Table~\ref{tab:training_budget} reports the real-simulator and imagined-rollout budgets used in each experiment. For Scenarios~1-3, direct model-free retuning and latent model-based retuning use the same scenario-specific real-simulator interaction budget; the model-based variants additionally reuse nominal latent structure learned before the abnormality is introduced, which is part of the proposed adaptation mechanism rather than an additional abnormal-scenario data source.

For Scenarios~1-3, each subsection discusses the scenario-level behavior, while Table~\ref{tab:abnormal_summary} reports the best final model-based variant against direct model-free retuning and Appendix~\ref{app:abnormal_tables} gives the full per-variant values.

\begin{table}[t]
	\centering
	\caption{ Simulator-interaction budgets used in the reported comparisons. Real steps denote high-fidelity simulator control steps; imagined steps are latent model rollouts and do not require simulator calls.}
	\label{tab:training_budget}
	\small
	\resizebox{\textwidth}{!}{%
	\begin{tabular}{llrrrrr}
		\toprule
		Study & Controller family & Epochs & Real steps/epoch & Total real steps & Imagined steps/epoch & Test episodes/epoch \\
		\midrule
		Model-free observability endpoints & Privileged / well-only SAC & 30 & 6000 & 180000 & -- & 1 \\
		Model-free deployable variants & History, masking, teacher-student SAC & 30 & 3000 & 90000 & -- & 1 \\
		Scenario~0 nominal model-based & Latent Dreamer-style control & 20 & 200 & 4000 & 4000 & 1 \\
		Scenarios~1--3 model-based retuning & Latent Dreamer-style retuning & 20 & 40 & 800 & 3200 & 1 \\
		Scenarios~1--3 model-free retuning & Direct SAC retuning & 20 & 40 & 800 & -- & 1 \\
		\bottomrule
	\end{tabular}%
	}
	
\end{table}

\subsection{Model-free results under varying observability}

\begin{figure}[t]
    \centering
    \includegraphics[width=0.9\linewidth]{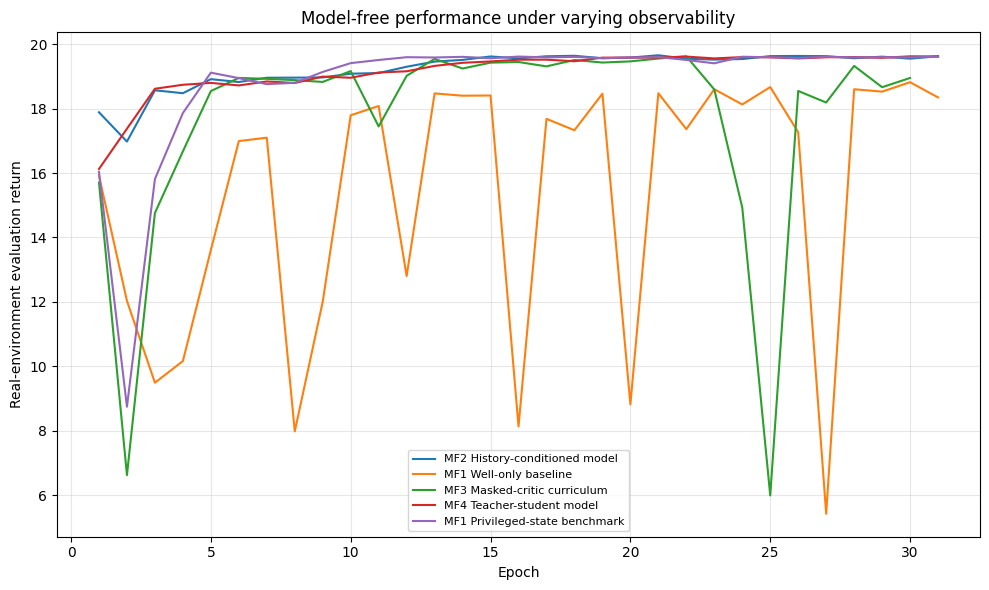}
    \caption{Test return as a function of training epoch for the five model-free variants. The well-only baseline is the weakest and most unstable controller, whereas the history-conditioned and teacher-student models reach the same high-performance regime as the privileged-state benchmark. The masking-curriculum variant achieves strong returns but is noticeably less stable.}
    \label{fig:mf_results}
\end{figure}

\begin{table}[t]
	\centering
	\caption{ Model-free controller performance under different observability regimes. Best and final values are evaluation returns from the TensorBoard logs.}
	\label{tab:model_free_results}
	\small
	\resizebox{\textwidth}{!}{%
	\begin{tabular}{llllrr}
		\toprule
		Method & Deployment input & Privileged training signal & Deployable & Best return & Final return \\
		\midrule
		Privileged-state benchmark & Spatial state + wells & Policy observes spatial state & No & 19.612 & 19.604 \\
		Well-only baseline & Current well tensor & None & Yes & 18.816 & 18.345 \\
		History-conditioned model & Well history + current wells & None & Yes & 19.652 & 19.622 \\
		Masked-critic curriculum & Current wells & Critic receives gradually masked spatial state & Yes & 19.619 & 18.948 \\
		Teacher-student model & Well history + current wells & Privileged teacher critics / distillation & Yes & 19.620 & 19.616 \\
		\bottomrule
	\end{tabular}%
	}
	
\end{table}

Figure~\ref{fig:mf_results} compares the test performance of the five model-free variants. The corresponding best and final evaluation returns are reported in Table~\ref{tab:model_free_results}. The well-only baseline is clearly the weakest and most unstable controller, confirming that decisions based only on the most recent well-response tensor are too information-limited for robust closed-loop control. In contrast, the privileged-state benchmark rapidly reaches high performance and remains stable, as expected when dense simulator fields are available to the policy.

The key result, however, is that the history-conditioned model closes almost all of the gap to the privileged benchmark while remaining fully deployable. Its performance rises steadily and stabilizes in the same high-return regime as the privileged model. This shows that much of the information lost under partial observability can be recovered from temporal structure in the public observation stream alone. The teacher-student model reaches essentially the same final performance band and does so with similarly smooth learning dynamics, indicating that asymmetric training with privileged critics can yield a deployable controller that remains competitive with the strongest history-based model.

The masking-curriculum variant provides a more mixed picture. It achieves strong returns for a substantial portion of training, which suggests that gradually withdrawing privileged critic information is a meaningful idea. However, it is noticeably less stable than the history-conditioned and teacher-student models, with pronounced late-training degradation. Taken together, these results support two main conclusions. First, temporal context is the dominant ingredient for overcoming partial observability in this setting. Second, privileged information is still useful during training, but its benefits are most convincing when incorporated through asymmetric teacher-student learning rather than through the masking curriculum alone.

\subsection{Scenario 0: nominal model-based retention}

\begin{figure}[t]
    \centering
    \includegraphics[width=0.9\linewidth]{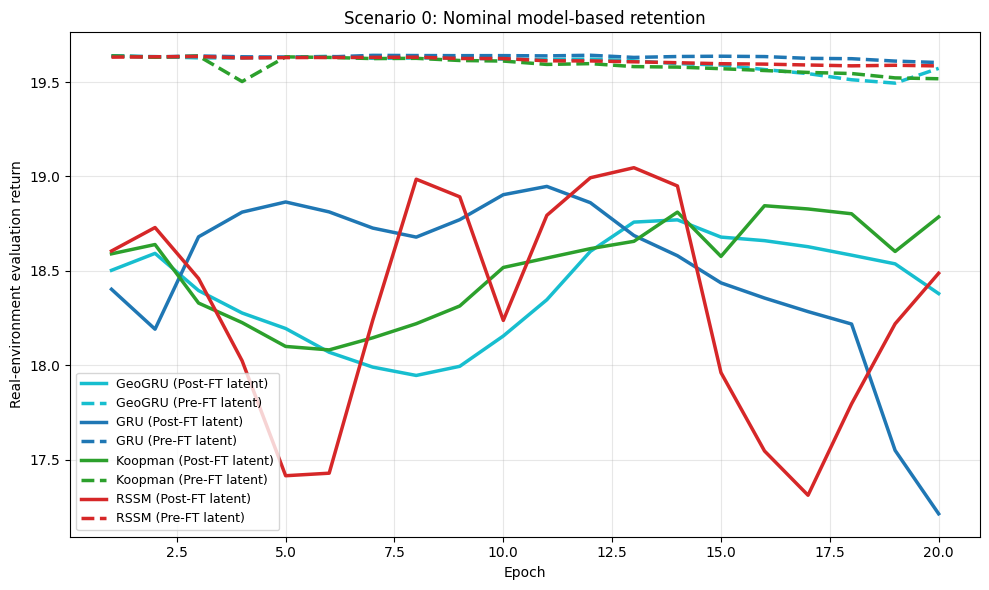}
    \caption{Scenario 0: nominal model-based retention. Real-environment evaluation return as a function of training epoch for the different world-model backbones and latent sources. Several model-based variants remain in the same high-performance regime as the deployable model-free reference, while clear differences also emerge between pre-finetuning and post-finetuning latents.}
    \label{fig:scenario0_results}
\end{figure}

\begin{table}[t]
	\centering
	\caption{Scenario~0 nominal model-based retention results. The deployable model-free teacher-student final return is 19.616; the retention column reports final return relative to this reference.}
	\label{tab:scenario0_results}
	\small
	\begin{tabular}{llrrr}
		\toprule
		World model & Latent source & Best return & Final return & Retention (\%) \\
		\midrule
		GRU & Pre-FT & 19.642 & 19.604 & 99.9 \\
		RSSM & Pre-FT & 19.636 & 19.586 & 99.8 \\
		GeoGRU & Pre-FT & 19.639 & 19.573 & 99.8 \\
		Koopman & Pre-FT & 19.639 & 19.518 & 99.5 \\
		Koopman & Post-FT & 18.845 & 18.786 & 95.8 \\
		RSSM & Post-FT & 19.047 & 18.487 & 94.2 \\
		GeoGRU & Post-FT & 18.770 & 18.379 & 93.7 \\
		GRU & Post-FT & 18.947 & 17.213 & 87.7 \\
		\bottomrule
	\end{tabular}
	
\end{table}

Figure~\ref{fig:scenario0_results} evaluates the model-based pipeline under nominal conditions. Table~\ref{tab:scenario0_results} reports the corresponding best and final returns and expresses the final values as retention relative to the deployable model-free teacher-student reference. The purpose of this experiment is not to outperform the model-free reference, but to determine whether latent imagination can preserve its real-environment performance. The results show that this is indeed possible. Several model-based variants remain in the same high-performance regime as the deployable model-free controller, demonstrating that the fixed latent model-based pipeline is viable in the base case.

A clear pattern also emerges with respect to the latent source. The controllers built on pre-finetuning latents are consistently stronger and more stable than most of their post-finetuning counterparts. In particular, the pre-finetuning variants remain tightly clustered near the top of the plot throughout training, whereas post-finetuning variants exhibit larger performance spread and, in some cases, marked degradation. This suggests that latent representations optimized for distillation or auxiliary objectives are not automatically the best representations for downstream model-based control. In other words, nominal control retention depends not only on the world-model backbone, but also on the structure of the latent space on which it is trained.

From the perspective of the paper, Scenario~0 therefore establishes the main prerequisite for the later abnormal cases: dream-based latent control can remain competitive in the real simulator under nominal physics, provided that the latent representation and world-model backbone are chosen appropriately.

\subsection{Scenario 1: adaptation under known injector failure}

\begin{figure}[t]
    \centering
    \includegraphics[width=0.9\linewidth]{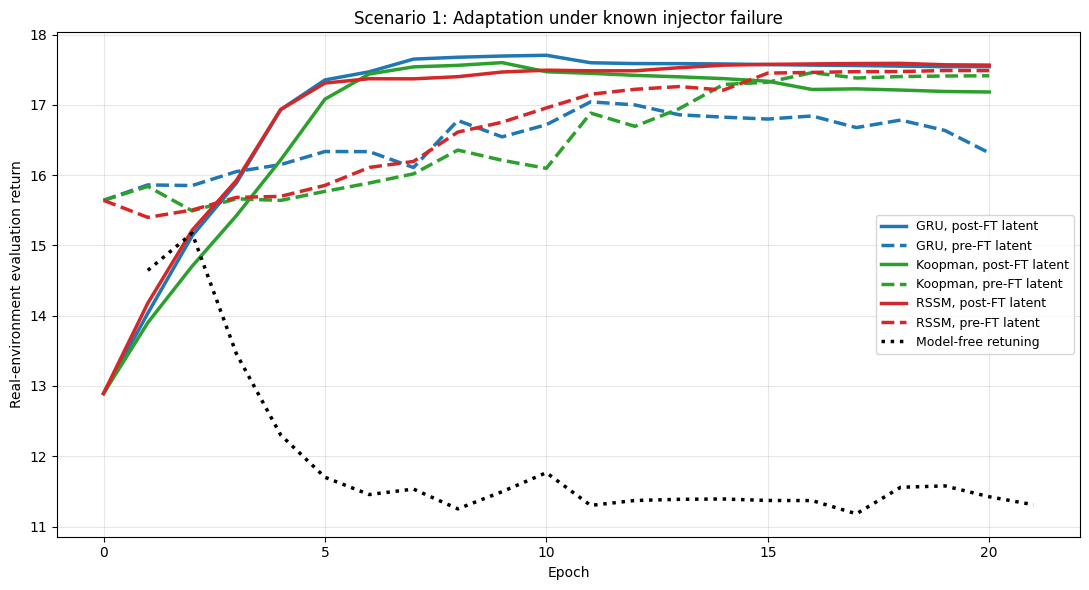}
    \caption{Scenario 1: adaptation under known injector failure. Real-environment evaluation return as a function of training epoch for model-based retuning and direct model-free retuning. Epoch 0 corresponds to the nominal controller before any scenario-specific updates. Model-based retuning rapidly recovers and clearly outperforms direct model-free retuning throughout training.}
    \label{fig:scenario1_results}
\end{figure}

Figure~\ref{fig:scenario1_results} shows the effect of losing one injector control channel. This is the easiest of the abnormal scenarios conceptually, since the abnormality is operational and explicitly known rather than latent and hidden. Even so, the difference between model-based and model-free adaptation is pronounced. After the shared nominal initialization at epoch 0, the model-based variants rapidly improve and converge to a high-performance regime, whereas direct model-free retuning deteriorates sharply and remains far below all model-based alternatives.

Thus, when the actuator loss is known and can be imposed consistently in real and imagined interaction, latent model-based retuning uses the limited abnormal simulator budget much more effectively than direct SAC retuning. The modest spread among model-based variants suggests that Scenario~1 mainly tests adaptation to a known control constraint rather than correction of deep latent-dynamics error.

\subsection{Scenario 2: adaptation under leakage-induced dynamics and reward shift}

\begin{figure}[t]
    \centering
    \includegraphics[width=0.9\linewidth]{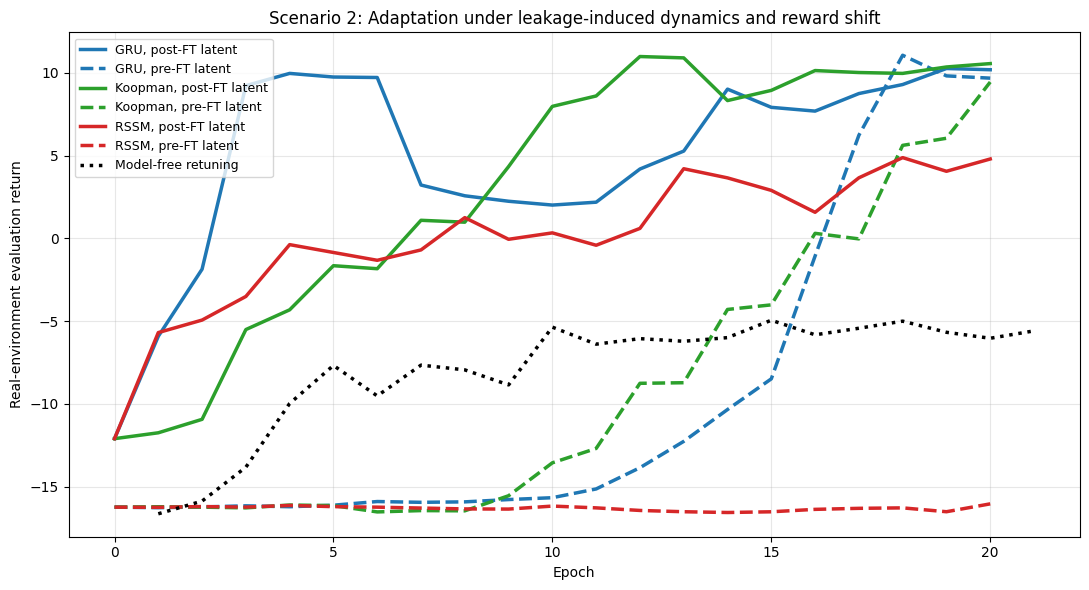}
    \caption{Scenario 2: adaptation under leakage-induced dynamics and reward shift. Real-environment evaluation return as a function of training epoch for model-based retuning and direct model-free retuning. Epoch 0 denotes the nominal controller before adaptation. The best model-based variants recover strongly despite the combined dynamics and reward shift, whereas model-free retuning remains in a poor regime.}
    \label{fig:scenario2_results}
\end{figure}

Figure~\ref{fig:scenario2_results} presents the most challenging and, arguably, most informative abnormal case. Here the nominal assumptions fail not only at the level of state evolution, but also at the level of the operational objective, since leakage introduces an additional penalty term. This is reflected in the much larger spread across model-based variants and the substantial difficulty of direct model-free retuning.

The best model-based variants recover to clearly positive high-return regimes, while direct model-free retuning remains poor throughout training. Unlike Scenario~1, however, performance depends strongly on the latent source and world-model backbone. This indicates that leakage adaptation is not only a policy-retuning problem; it also tests whether the residual adapter can correct a meaningful dynamics-and-reward mismatch.

\subsection{Scenario 3: adaptation under compartmentalized connectivity shift}

\begin{figure}[t]
    \centering
    \includegraphics[width=0.9\linewidth]{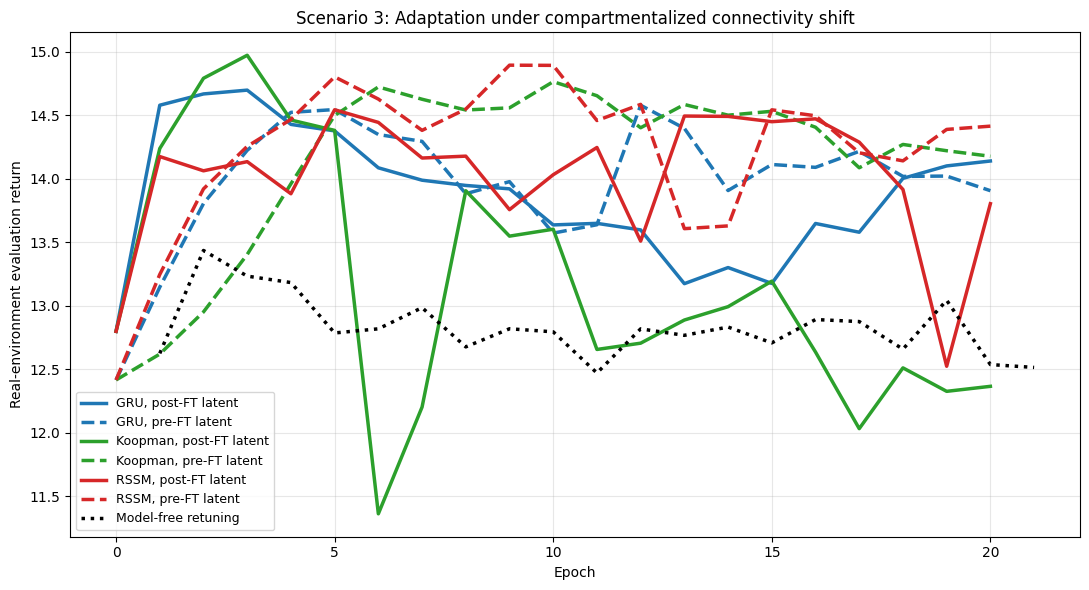}
    \caption{Scenario 3: adaptation under compartmentalized connectivity shift. Real-environment evaluation return as a function of training epoch for model-based retuning and direct model-free retuning. Epoch 0 denotes the nominal controller before adaptation. Most model-based variants outperform model-free retuning, although the spread across latent sources and world-model backbones is more nuanced than in Scenario 2.}
    \label{fig:scenario3_results}
\end{figure}

Figure~\ref{fig:scenario3_results} considers a different kind of physics mismatch. Specifically, the reservoir is more compartmentalized than assumed in the nominal digital twins. Unlike Scenario~2, the reward structure is unchanged, so the challenge lies primarily in the transition dynamics rather than in both dynamics and objective. The results again favor model-based adaptation, although the margin over model-free retuning is smaller and more nuanced than in the leakage scenario.

Most model-based variants improve substantially over the nominal initialization and remain above the model-free retuning curve for the majority of training. However, the scenario is clearly more heterogeneous than Scenario~1, and the relative ranking of world-model/latent combinations is less uniform. Some variants remain strong and stable, while others display noticeable fluctuations or partial degradation. This indicates that the compartmentalized setting is difficult in a different way from leakage: the problem is not an explicit new penalty, but a structural revision of flow connectivity that must be absorbed by the adapted latent dynamics.

The main conclusion is nevertheless consistent with the earlier abnormal scenarios. Even when the control interface is unchanged and the reward is nominal, correcting the nominal latent dynamics through the model-based pipeline remains more effective than direct model-free retuning. The advantage is smaller than in Scenario~2, but it is still systematic enough to support the broader adaptation argument of the paper.

\begin{table}[t]
	\centering
	\caption{Abnormal-scenario retuning summary. For each scenario, the model-based row reports the best final-performing model-based variant. The last column is the final-return difference relative to direct model-free retuning under the same scenario-specific real-simulator budget.}
	\label{tab:abnormal_summary}
	\small
	\resizebox{\textwidth}{!}{%
	\begin{tabular}{lllrrrr}
		\toprule
		Scenario & Abnormality & Best final MB variant & MF retuning final & MB best return & MB final return & Final gain \\
		\midrule
		1 & Known injector-control loss & GRU Post-FT & 11.573 & 17.725 & 17.682 & +6.109 \\
		2 & Leakage dynamics + reward shift & Koopman Post-FT & -5.584 & 10.990 & 10.563 & +16.147 \\
		3 & Compartmentalized connectivity shift & RSSM Pre-FT & 12.514 & 14.894 & 14.414 & +1.900 \\
		\bottomrule
	\end{tabular}%
	}
	
\end{table}

\subsection{Overall interpretation of the results}

Across the abnormal scenarios, the benefit of model-based retuning is largest when the mismatch changes the reward and dynamics simultaneously, as in leakage, and smaller when the mismatch is primarily structural, as in compartmentalization. The results also show that no single world-model backbone or latent source dominates uniformly. In particular, the contrast between pre-finetuning and post-finetuning latent representations suggests that latent design is part of the adaptation problem itself, rather than a purely technical implementation detail.

\section{Conclusions}

This work formulated closed-loop CO$_2$ storage control as a partially observable sequential decision problem and evaluated deployable model-free and latent model-based controllers under nominal and abnormal operating conditions. The results show that well-history conditioning is sufficient to recover nearly all of the privileged-state performance while preserving deployment feasibility. They also show that latent model-based adaptation can use limited abnormal-scenario simulator interaction more effectively than direct model-free retuning, especially when the mismatch alters both dynamics and reward.

A broader conclusion is that latent representation design matters throughout the workflow. The comparison between pre-finetuning and post-finetuning latents shows that the latent space learned for model-free control is not neutral with respect to downstream model-based control and adaptation. Similarly, no single world-model backbone dominated uniformly across all scenarios, which suggests that robustness in this setting emerges from the interaction between encoder, latent geometry, and adaptation mechanism rather than from a single universally best model class.

Overall, the results support a practical message for CCS closed-loop control. Strong deployable policies can be learned from public well-level information when temporal context is modeled explicitly, and these policies can later be adapted effectively through latent model-based pipelines when the control interface or the reservoir physics deviate from nominal assumptions. This makes the proposed framework a viable alternative to repeated history matching and re-optimization in settings where online adaptation must remain computationally tractable.

The present results should be interpreted as simulator-budget-controlled comparisons rather than exhaustive robustness studies. Future work should therefore examine multi-seed variability, richer monitoring streams, online updating of the public encoder, and abnormal scenarios in which the type or location of the mismatch is not known in advance. Additional extensions include latent representations explicitly co-designed for both model-free control and downstream model-based adaptation, more realistic abnormality detection and partial diagnosis settings, and richer closed-loop decision problems involving structural uncertainty updates, field-development actions, and multi-stage operational planning. These directions would further strengthen the connection between deployable deep reinforcement learning and practical CCS reservoir management.

 \section*{Credit author statement}
 Conceptualization, S.F.; methodology, S.F.; software, S.F.; validation, S.F. and V.G.; resources, S.F.; writing---original draft preparation, S.F.; writing---review and editing, S.F., V.G.; visualization, S.F. All authors have read and agreed to the final version of the manuscript.

\section*{Declaration of competing interest} The authors declare that they have no known competing financial interests or personal relationships that could have appeared to influence the work reported in this paper.

\section*{Funding} The resources were granted with the support of GRNET as part of the project StorageCO$_2$.
The publication of the article in OA mode was financially supported by HEAL-Link.

\section*{Data Availability Statement} Project code can be found at https://github.com/flammmes/Closed-Loop-Reservoir-Management

\newpage

\appendix

\section{Detailed abnormal-scenario retuning results}
\label{app:abnormal_tables}

This appendix reports the full per-variant abnormal-scenario retuning values that support the scenario-level summary in Table~\ref{tab:abnormal_summary}. These tables are kept outside the main text because the main claims depend on the scenario-level comparison between direct model-free retuning and the best final-performing model-based retuning variant under the same scenario-specific real-simulator budget.

\begin{table}[htbp]
	\centering
	\caption{ Full Scenario~1 retuning results under known injector-control loss.}
	\label{tab:scenario1_full}
	\small
	\begin{tabular}{llrr}
		\toprule
		Run & Family & Best return & Final return \\
		\midrule
		MB GRU Post-FT & Model-based & 17.725 & 17.682 \\
		MB Koopman Post-FT & Model-based & 17.683 & 17.242 \\
		MB RSSM Post-FT & Model-based & 17.578 & 17.500 \\
		MB Koopman Pre-FT & Model-based & 17.492 & 17.465 \\
		MB RSSM Pre-FT & Model-based & 17.471 & 17.319 \\
		MB GRU Pre-FT & Model-based & 17.328 & 17.164 \\
		MF retuning & Model-free & 14.645 & 11.573 \\
		\bottomrule
	\end{tabular}
	
\end{table}

\begin{table}[htbp]
	\centering
	\caption{ Full Scenario~2 retuning results under leakage-induced dynamics and reward shift. Runs with missing evaluation values are omitted from the table.}
	\label{tab:scenario2_full}
	\small
	\begin{tabular}{llrr}
		\toprule
		Run & Family & Best return & Final return \\
		\midrule
		MB GRU Pre-FT & Model-based & 11.066 & 9.683 \\
		MB Koopman Post-FT & Model-based & 10.990 & 10.563 \\
		MB GRU Post-FT & Model-based & 10.266 & 10.192 \\
		MB Koopman Pre-FT & Model-based & 9.449 & 9.449 \\
		MB RSSM Post-FT & Model-based & 4.883 & 4.800 \\
		MB RSSM Pre-FT & Model-based & -16.034 & -16.034 \\
		MF retuning & Model-free & -4.941 & -5.584 \\
		\bottomrule
	\end{tabular}
	
\end{table}

\begin{table}[htbp]
	\centering
	\caption{ Full Scenario~3 retuning results under compartmentalized connectivity shift.}
	\label{tab:scenario3_full}
	\small
	\begin{tabular}{llrr}
		\toprule
		Run & Family & Best return & Final return \\
		\midrule
		MB Koopman Post-FT & Model-based & 14.972 & 12.365 \\
		MB RSSM Pre-FT & Model-based & 14.894 & 14.414 \\
		MB Koopman Pre-FT & Model-based & 14.764 & 14.177 \\
		MB GRU Post-FT & Model-based & 14.697 & 14.139 \\
		MB GRU Pre-FT & Model-based & 14.578 & 13.906 \\
		MB RSSM Post-FT & Model-based & 14.543 & 13.803 \\
		MF retuning & Model-free & 13.435 & 12.514 \\
		\bottomrule
	\end{tabular}
	
\end{table}


\bibliographystyle{unsrt} 

\bibliography{references.bib}

\end{document}